\documentclass[review]{elsarticle}
\usepackage{bbm,amsmath,amssymb,upgreek,lscape,mathrsfs}
\usepackage{subfig}
\usepackage{cancel}
\usepackage{epsfig}
\usepackage{diagbox}
\usepackage[a4paper, total={6in, 8in}]{geometry}
    
\usepackage[skins,theorems]{tcolorbox}
\tcbset{highlight math style={enhanced,
  colframe=black,colback=white,arc=0pt,boxrule=1pt}}
\usepackage{lineno,hyperref}
\modulolinenumbers[5]
\usepackage{xcolor}
\journal{Journal of \LaTeX\ Templates}
\graphicspath{{Figures/}}

\begin{document}

\begin{frontmatter}

\title{Energy-Time Optimal Control of Wheeled Mobile Robots}

\author{\fnref{myfootnote} Youngjin Kim }
\fntext[myfootnote]{Dept. of Mechanical and Aerospace Engineering, University at Buffalo (SUNY), Buffalo, NY, 14260-4400, USA}

\author[mysecondaryaddress]{Tarunraj Singh\corref{mycorrespondingauthor}}
\cortext[mycorrespondingauthor]{Corresponding author}
\ead{tsingh@buffalo.edu}


\begin{abstract}
	    This paper focuses on the energy-time optimal control of wheeled mobile robots undergoing point-to-point transitions in an obstacles free space. Two interchangeable models are used to arrive at the necessary conditions for optimality. The first formulation exploits the Hamiltonian, while the second formulation considers the first variation of the augmented cost to derive the necessary conditions for optimality. Jacobi elliptic functions are shown to parameterize the closed form solutions for the states, control and costates.
	Analysis of the optimal control reveal that they are constrained to lie on a cylinder whose circular cross-section is a function of the weight penalizing the relative costs of time and energy.
	The evolving optimal costates for the second formulation are shown to lie on the intersection of two cylinders. The optimal control for the wheeled mobile robot undergoing point-to-point motion is also developed where the linear velocity is constrained to be time-invariant. It is shown that the costates are constrained to lie on the intersection of a cylinder and an extruded parabola. Numerical results for various point-to-point maneuvers are presented to illustrate the change in the structure of the optimal trajectories as a function of the relative location of the terminal and initial states. 
\end{abstract}


\end{frontmatter}

\nolinenumbers

\section{Introduction}
Optimal control in numerous applications seeks to minimize multiple objectives~\cite{nabawy2018energy, faraj2016optimal, hindle2001robust}. One such combination of costs include maneuver time and energy consumed. It has been shown that jumping spiders minimize flight time for short range jumps while they optimize for energy consumed for longer jumps~\cite{nabawy2018energy}. Finite battery capacity limit the range of electric vehicles and it is easy to motivate the need for energy optimal control~\cite{faraj2016optimal}. However, electric vehicle are bound by road speed constraints and occasionally need to speed up to overtake vehicles calling for minimum time maneuvers. Karri et al.~\cite{2020-01-0579} study connected and automated vehicles where they consider the issue of minimizing energy consumed by the vehicle while minimizing travel time as well. They experimentally demonstrate their control on an Audi A3 e-tron, in complex driving scenarios which include on-ramp merging, round-about conflict resolution and speed-reduction zones.
    Similarly, finite fuel on spacecrafts also motivate a need for energy-time optimal control~\cite{liu1997fuel}. In this paper, energy-time optimal solutions for wheeled mobile robot (WMR) which are battery driven are considered for path planning in an environment without obstacles. The motivation for minimum energy cost is justified given the fact that battery driven mobile robots' endurance can be extended by judicious controllers which minimize consumed energy. However, one needs to specify the maneuver time which can be arbitrary. A weighted time/energy cost function is considered in this work where the user selected weight permits the optimal control to range from a time-optimal to a minimum energy solution. 

Wheeled mobile robots have been extensively studied motivated by their potential applications in floor cleaning~\cite{palacin2004building}, personal transportation~\cite{pinto2012development}, lawn mowing~\cite{hicks2000survey}, and have been used as rudimentary models for cars and trailers~\cite{murray1990steering}. They have also served as particle models for formation control of multi-agent systems~\cite{justh2015optimality}. 
Klancar et al.~\cite{klancar2017wheeled} provide a concise introduction to wheeled mobile robots including aspects of modeling, estimation, path planning and control. They present examples of problems including the differential drive robot with two independently controlled wheels and bicycle models with front wheel steering, which are modelled by what is popularly referred to as the unicycle model. 

Basic kinematic models for wheeled mobile robots have been popular models for controller design~\cite{justh2015optimality, maclean2013path, mukherjee1997optimal}, path planning~\cite{pathak2005integrated} and multiple interacting agents (vehicles)~\cite{paden2016survey}. The two input, three state model which goes by the moniker ``Unicycle model'' was studied by Dubins~\cite{dubins1957curves} to determine the shortest path of a particle moving with constant velocity between two points given constraints on the curvature of the path. Reeds and Shepp~\cite{reeds1990optimal} extended the Dubins' solution to vehicles which can moves forward and backward. 
{Liu and Sun~\cite{liu2013minimizing} considered the problem of energy optimal motion planning for wheeled mobile robots where they parameterized the position and orientation states using cubic Bezier curves, while Sathiya and Chinnadurai~\cite{sathiya2019evolutionary} use NURBS to parameterize the trajectories to determine trajectories which minimze a multi-objective cost function which include maneuver time and actuator efforts. In this work, no such parameterization of the trajectories is assumed since we demonstrate that Jacobi elliptic functions are the functions which exactly solve the control problem.} 
Minimum energy controllers for the unicycle model for point to point path planning have also been studied~\cite{justh2015optimality,  maclean2013path, mukherjee1997optimal, kim2021minimum, butt2015integrability}, where they demonstrated that the optimal paths were parameterized using Jacobi elliptic functions. Justh and Krishnaprasad~\cite{justh2015optimality} studied the  minimum energy optimal control problem for the Dubins vehicle (i.e., where the linear velocity in unity). They formulated an optimal control problem for systems defined on a Lie group. They pose a cost function for the particle collective that concurrently  penalizes the individual control magnitude and the difference among the individual controls. Their solution suggests an optimality principle for collective strategy in animal behavior called {\it Allelomimesis}.
Maclean and Biggs~\cite{maclean2013path} considered two problems: (1) one where the linear and angular velocities are arbitrary and (2) one where the linear velocity is unity and studied the minimum energy problem using the machinery of Lie groups. {Halder and et al.~\cite{Halder2017TimeOptimalSF} used geometric control theory on the special Euclidean group SE(2) to minimize steering control effort and the solutions were found by numerical approach . Similar study was introduced ~\cite{Tiwari.2021}, where they consider an energy optimal cost where the final time in prescribed. The authors claim that they cannot develop an analytical solution to the optimal control problem and explore the use of exotic integrators to solve the problem numerically.}

Mukherjee et al.~\cite{mukherjee1997optimal} derived the necessary conditions for optimality for the minimum energy problem and numerically solved the  two  point  boundary  value  problem  using  a  relaxation method. They noted the optimal motion of the differential wheeled robot was similar to the motion of a pendulum in a gravitational field.  They conclude that the optimal trajectories can also be determined by solving for four constants constrained  by  four  nonlinear  equations  which  result  from the analytical solution to the pendulum motion in a gravity field. It should be noted that the papers cited include either minimum energy or minimum time control where the linear velocity is fixed. This paper focuses on a energy-time optimal problem which subsumes the energy or time-optimal control problems in a unified framework. {The fundamental contributions of this work are: (1) Two distinct optimal control formulations for two different representations of the kinematics of wheeled mobile robots which provide a unique perspective on the constraints on the control and costates, (2) Closed form solutions to the optimal states, costates and control and, (3) conversion of the optimal control problem to a nonlinear programming problem which permits real-time computation of the optimal paths for point-to-point maneuvers and (4) Closed form solutions to a constant linear velocity problem which minimize the weighted energy/time cost.}

This paper begins by reviewing two representations of the kinematic model of  a wheeled mobile robot. This is followed in Section 2 with the detailed development of the necessary conditions for optimality for the minimum energy-time cost function for the two kinematic models. Analytical solutions for the states, control and costates are derived for both formulations. Section 3 presents numerical results for a slew of terminal conditions to illustrate the interesting transitions in the control profiles and trajectories. Section 4 presents the energy-time optimal development for the Dubins' problem where the linear velocity is time-invariant and numerical results are presented to illustrate the optimal trajectories. The paper concludes with a summary of the results of the paper.

\section{Problem Formulation}

\begin{figure}[!h]     
\centering
    \includegraphics[width=0.5\textwidth]{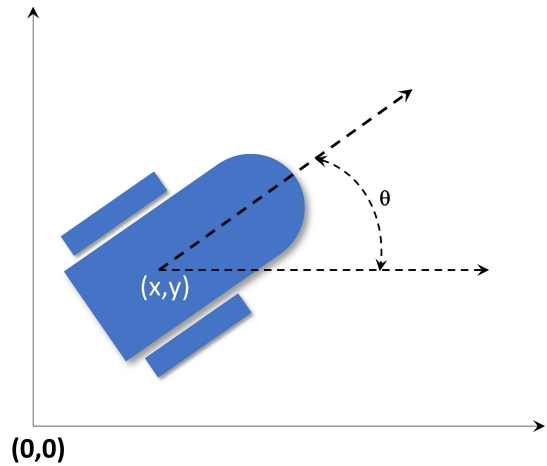}
    \caption{Wheeled Mobile Robot} 
    \label{fig:WMR}
\end{figure}
A popular model for a differentially driven robot moving with a velocity $v$ along the vector given by the orientation $\theta$ is modelled by the kinematics:
\begin{eqnarray}
    \cos(\theta) \dot{x} + \sin(\theta) \dot{y} & =  v \label{eq:1WMB}\\
    -\sin(\theta) \dot{x} + \cos(\theta) \dot{y} & =  0 \label{eq:2WMB} \\
    \dot{\theta} & =  \omega. \label{eq:3WMB}
\end{eqnarray}
as shown in Figure~\ref{fig:WMR}.
Equation~\ref{eq:2WMB} is a non-holonomic constraint which enforces the requirement that the instantaneous lateral velocity of the wheeled mobile robot be zero. A passive transformation of the coordinate reference frame through the orientation angle $\theta$, leads to 
Equation~\ref{eq:1WMB}-\ref{eq:3WMB} being rewritten as:
\begin{equation}
  \underset{\mathbf{\dot{X}}}{\underbrace{\begin{Bmatrix}  \dot{x} \\ \dot{y} \\  \dot{\theta} \end{Bmatrix}}} = \underset{f(\mathbf{X},\mathbf{U})}{\underbrace{\begin{Bmatrix} \cos(\theta) v \\ \sin(\theta) v \\ \omega \end{Bmatrix}}}, \label{eq:veceq}
\end{equation}
where $\mathbf{X}$ is the state vector and $\mathbf{U}$ is the control vector.

The problem considered in this work is that of minimizing a weighted combination of energy and maneuver time:
\begin{equation}
    J = \int_0^{T_f} \left( (1-\mu) + \frac{\mu}{2} \left( v^2 + \omega^2 \right) \right) \: dt
\end{equation}
subject to transitioning the system states from:
\begin{equation}
    x(0) = 0, y(0) = 0, \theta(0) = 0
\end{equation}
to: 
\begin{equation}
    x(T_f) = x_f, y(T_f) = y_f,
\end{equation}
allowing the terminal orientation of the robot to be {free}.
 The weighting parameter $\mu$ lies between 0 and 1 and permits the solution to transition between a time optimal ($\mu=0$) to an energy optimal ($\mu=1$) solution. Parallel developments of the optimal control problem for the two system models given by Equation~\ref{eq:1WMB}-\ref{eq:3WMB} and \ref{eq:veceq} will be presented. Equation~\ref{eq:veceq} permits the use of the Hamiltonian to arrive at the necessary conditions for optimality, while the necessary conditions for optimality for the model given by Equation~\ref{eq:1WMB}-\ref{eq:3WMB} are derived from variational calculus. Since the maneuver time $T_f$ is free, a new normalized time is defined as $\tau = \frac{t}{T_f}$ which transforms the cost function to:
\begin{equation}
    J = T_f \int_0^{1} \left( (1-\mu) + \frac{\mu}{2} \left( v^2 + \omega^2 \right) \right) \: d\tau.
    \label{eq:costfn_tau}
\end{equation}
This cost function in conjunction with the system dynamics transformed into the $\tau$ space will be used to derive the optimal control profiles.

\subsection{First Formulation} \label{sec:Form1}
The Hamiltonian in the non-dimensional time $\tau$ is defined as:
\begin{equation}
    \mathcal{H} = T_f (1-\mu)+T_f\mu\frac{\omega^2}{2}+T_f\mu\frac{v^2}{2}+\lambda_1T_fv\cos(\theta)+\lambda_2T_fv\sin(\theta)+\lambda_3T_f\omega, \label{eq:Hameq}
\end{equation}
where a new state $T_f$ is introduced whose rate of change is zero. 
The Hamiltonian leads to the necessary conditions for optimality:
 
\begin{align}
\mathbf{\dot{X}} & = \frac{\partial \mathcal{H}}{\partial \boldsymbol{\lambda}}  \label{eq:Hst} \\
\boldsymbol{\dot{\lambda}} & = -\frac{\partial \mathcal{H}}{\partial \mathbf{X}} \label{eq:Hcst} \\
\frac{\partial \mathcal{H}}{\partial \mathbf{U}} & = 0 \label{eq:Hctr}
\end{align}
when the control inputs $v$ and $\omega$ are unconstrained.

The coupled state and costate equations are given by:
\begin{align}
           \dot{x}&=\frac{\partial \mathcal{H}}{\partial \lambda_1}=T_f v\cos(\theta) \\
           \dot{y}&=\frac{\partial \mathcal{H}}{\partial \lambda_2} =T_f v\sin(\theta) \\
           \dot{\theta}&= \frac{\partial \mathcal{H}}{\partial \lambda_3} =T_f\omega \label{eq:tdor_form1}\\
           \dot{T_f}&= \frac{\partial \mathcal{H}}{\partial \lambda_4} =0\\
           \dot{\lambda_1}&=-\frac{\partial \mathcal{H}}{\partial x}= 0, \quad  \Rightarrow \lambda_1 = C_1 \label{eq:lam1}\\
           \dot{\lambda_2}&=-\frac{\partial \mathcal{H}}{\partial y}=0, \quad  \Rightarrow \lambda_2 = C_2 \label{eq:lam2}\\
           \dot{\lambda_3}&=-\frac{\partial \mathcal{H}}{\partial \theta}= T_f\lambda_1v\sin(\theta)-T_f\lambda_2v\cos(\theta) \label{eq:lam3}\\
           \dot{\lambda_4}&=-\frac{\partial \mathcal{H}}{\partial T_f}=  -(1-\mu)-\mu\frac{\omega^2}{2}-\mu\frac{v^2}{2}-\lambda_1v\cos(\theta)-\lambda_2v\sin(\theta)-\lambda_3\omega \label{eq:L4dot}
\end{align}
with the optimal control given by the equations:

\begin{align}
           \frac{\partial \mathcal{H}}{\partial v} &=\mu T_fv+\lambda_1T_f\cos(\theta)+\lambda_2T_f\sin(\theta)= 0 \\
           \Rightarrow v &= \frac{-\lambda_1\cos(\theta)-\lambda_2\sin(\theta)}{\mu} \label{eq:optv} \\
           \frac{\partial \mathcal{H}}{\partial \omega} &= \mu T_f\omega+\lambda_3T_f= 0 \\
           \Rightarrow \omega &= \frac{-\lambda_3}{\mu} \label{eq:optomg}
\end{align}
and the transversality conditions (TC) are:
\begin{align}
   x(0) &= 0, & y(0) &= 0, & \theta(0) &= 0,  \\
    x(1) &= x_f, & y(1) &= y_f, & &\\
    \lambda_3(1) &= 0, & \lambda_4(0) &= \lambda_4(1). & & \label{eq:Tc3}
\end{align}
{Please note that the terminal orientation of the robot is free, which leads to associated costate $\lambda_3$ at $\tau =1$ to be zero as shown in the equation \ref{eq:Tc3}. If the terminal orientation is explicitly assigned $\lambda_3$ at the terminal time will be free.}

The total time derivative of the Hamiltonian is:
\begin{equation}
\frac{d\mathcal{H}}{d\tau} = \frac{\partial \mathcal{H}}{dT_f} \cancelto{0}{\frac{dT_f}{d\tau}} + \frac{\partial \mathcal{H}}{d\mathbf{X}} \frac{d\mathbf{X}}{d\tau} + \frac{\partial \mathcal{H}}{d\boldsymbol{\lambda}} \frac{d\boldsymbol{\lambda}}{d\tau}+ \cancelto{0}{\frac{\partial \mathcal{H}}{d\mathbf{U}}} \frac{d\mathbf{U}}{d\tau} = 0
\end{equation}
since the middle two terms cancel out due to Equations~\ref{eq:Hst}-\ref{eq:Hcst}.

Note that the Hamiltonian given by Equation~\ref{eq:Hameq} can be rewritten using Equation~\ref{eq:L4dot} as:
\begin{equation}
    \mathcal{H} = -T_f\dot{\lambda}_4.
\end{equation}
Since $\frac{d\mathcal{H}}{d\tau} = 0$, we have
\begin{equation}
    \mathcal{\dot{H}} = -\cancelto{0}{\dot{T}_f}\dot{\lambda}_4 - T_f\ddot{\lambda}_4 = 0,
\end{equation}
which requires $\ddot{\lambda}_4 = 0$ or $\dot{\lambda}_4$ is a constant. Since we require  $\lambda_4(0) = \lambda_4(1)$, this is only possible when $\dot{\lambda}_4(0) = 0$ which results in $\mathcal{H} = 0$ for all time.

Substituting Equation~\ref{eq:optv} and \ref{eq:optomg} into the Hamiltonian, reduces it to:
\begin{equation}
    \mathcal{H} = T_f\left(  (1-\mu)+\mu \frac{\omega^2}{2}+\mu \frac{v^2}{2}+\mu (-v^2-\omega^2)\right) = T_f\left((1-\mu)-\mu \frac{\omega^2}{2}-\mu \frac{v^2}{2}\right) = 0,
    \label{eq:Ham_constant}
\end{equation}
 which implies that the control for all time has to lie on a circle given by the equation: 
 \begin{equation}\tcbhighmath[drop fuzzy shadow]{v^2(\tau) + \omega^2(\tau) =   \left( \frac{\sqrt{2\mu(1-\mu)}}{\mu} \right)^2,}
 \label{eq:concons}
 \end{equation}   
  the radius of which $\frac{\sqrt{2\mu(1-\mu)}}{\mu}$ is only a function of the weighting parameter $\mu$ and does not depend on the terminal constraints. {Since the maximum control magnitude can be used to determine the corresponding limiting value of $\mu$ from Equation~\ref{eq:concons}, it can serve to determine the range of values of $\mu$ which can be used in the optimal control formulation.}
  
 The optimal control given by Equation~\ref{eq:optv} leads to:
    \begin{equation} \label{eq_v}
    \begin{split}
    v &= \frac{-C_1\cos(\theta)-C_2\sin(\theta)}{\mu} = \frac{-z}{\mu}\sin(\theta+\phi)
    \end{split}
    \end{equation}
where $\phi = \tan^{-1}(\frac{C_1}{C_2})$, $z = \sqrt{C_1^2+C_2^2}$,  $\lambda_1(t)=C_1$ and $\lambda_2(t)=C_2$ are constants. Further, Equation~\ref{eq:optomg} leads to:
    \begin{equation} \label{eq_w}
    \omega = \frac{-\lambda_3}{\mu}, 
    \Rightarrow \dot{\omega} = \frac{-\dot{\lambda}_3}{\mu}
    \end{equation}
which results in the derivative of Equation~\ref{eq:tdor_form1} reducing to:   
    \begin{equation} \label{eq_thddotN}
        \ddot{\theta} =  \dot{T}_f\omega+T_f\dot{\omega} =  T_f\dot{\omega}  = T_f\frac{-\dot{\lambda}_3}{\mu},
    \end{equation}
since $\dot{T}_f = 0 $.
%
Rewriting Equation~\ref{eq:lam3} as:    
\begin{equation} \label{eq_lam3}
\begin{split}
    \dot{\lambda}_3 & = T_fC_1v\sin(\theta)-T_fC_2v\cos(\theta) = 
-T_fvz\cos(\theta+\phi)
\end{split}
\end{equation}
and substituting into Equation~\ref{eq_thddotN}, we have:
    \begin{equation}
        \ddot{\theta} = T_f\frac{T_fvz\cos(\theta+\phi)}{\mu}=  -\frac{T_f^2z^2}{2\mu^2}\sin(2\theta+2\phi)
    \end{equation}
which can be solved as:    
\begin{equation}\label{eq:thdot_diff}
\begin{split}
    & \frac{1}{2}\dot{\theta}^2 = \frac{T_f^2z^2}{4\mu^2}\Big(\cos(2\theta+2\phi)+C_\theta\Big)= \frac{T_f^2z^2}{4\mu^2}\Big(Q- 2\sin^2(\theta+\phi)\Big) 
                \end{split}
\end{equation}    
where $Q = 1+C_\theta$, which leads to the closed form solution (details provided in the Appendix A, Equation ~\ref{eq:theta^2}-\ref{eq_apx:CF_theta}):
\begin{equation}
\theta (\tau) = \arcsin\Big(\sqrt{\frac{Q}{2}}\text{sn}(\frac{T_fz}{\mu}\tau+\eta,\frac{Q}{2})\Big)-\phi
\label{eq:theclf}
\end{equation}
where $\text{sn}(.,m)$ is the elliptic sine function which is the counterpart to the $\sin(.)$ trigonometric function. The Jacobi elliptic functions can be considered as a generalization to an ellipse of the trigonometric functions which are defined with reference to a circle. The Jacobi elliptic functions are parameterized in terms of the modulus $m$ which transitions the Jacobi elliptic function to trigonometric functions when $m$ = 0 and to hyperbolic functions when $m=1$. 

Substituting the initial condition $\theta(0) = 0$, leads to the closed form solution:
\begin{equation}\tcbhighmath[drop fuzzy shadow]{\phi = \arcsin\Big(\sqrt{\frac{Q}{2}}\text{sn}(\eta,\frac{Q}{2})\Big).}
\label{eq:phiclf}
 \end{equation}   
 
Closed form solution for the state $x(\tau)$ can also be shown to be (Equation \ref{eq:xdot2}, \ref{eq:cf_x}):
\begin{equation}
\begin{split}
           \quad x(\tau)& = \sqrt{m}\cos(\phi)\text{cn}(u, m)-\sin(\phi)(u-E(\text{amp}(u,m),m))+C_x
\end{split}
\label{eq:const1}
\end{equation}
where  $u = \frac{T_fz}{\alpha}\tau+\eta$, $m = \frac{Q}{2}$, $\text{cn}(.,m)$ is the elliptic cosine function, $E(u,m)$ is the incomplete
elliptic integral of the second kind, and $\text{amp}$ is the Jacobi amplitude function. Similarly, the state $y(\tau)$ (Equation \ref{eq:ydot2}, \ref{eq:cf_y}) is:
\begin{equation}
\begin{split}
           \quad y(\tau)& = -\cos(\phi)(u-E(\text{amp}(u,m),m))-\sqrt{m}\sin(\phi)\text{cn}(u, m)+C_y.
\end{split}
\label{eq:const2}
\end{equation}

From the initial condition $x(0)=0$, we have:
\begin{align*}
     C_x & = -\sqrt{m}\cos(\phi)\text{cn}(\eta, m) +\sin(\phi)(\eta-E(\text{amp}(\eta,m),m))
\end{align*}
and from initial condition $y(0)=0$:
\begin{align*}
          C_y & =\sqrt{m}\sin(\phi)\text{cn}(\eta, m)+ \cos(\phi)(\eta-E(\text{amp}(\eta,m),m)). 
\end{align*}

The third costate $\lambda_3(\tau)$ (Equations~\ref{eq:lam3_dot}, \ref{eq:cf_lam3}) is given as:
\begin{equation}
\begin{split}
\lambda_3(\tau) & = -z\sqrt{m}\text{cn}(u,m) +C_{\lambda_3}. \label{eq:lam3closed}
\end{split}
\end{equation}
Defining $\psi = \theta + \phi$, where $\phi$ is a constant, Equation~\ref{eq:thdot_diff} can be rewritten as:
\begin{equation}
                \frac{1}{2}\dot{\theta}^2 = \frac{1}{2}\dot{\psi}^2 =  \frac{T_f^2z^2}{4\mu^2}\Big(Q- 2\sin^2(\psi)\Big) = \frac{1}{2}\left( \frac{-T_f \lambda_3}{\mu} \right)^2.
                \label{eq:clam3}
\end{equation} 
Equation~\ref{eq:theclf} permits us to write:
\begin{equation}
    \frac{2}{Q} \sin^2(\psi) = \text{sn}^2(\frac{T_fz}{\mu}\tau+\eta,\frac{Q}{2}).
    \label{eq:coneqn5}
\end{equation}
Substituting Equation~\ref{eq:lam3closed} in Equation~\ref{eq:clam3} and using Equation~\ref{eq:coneqn5} leads to:
\begin{equation}
 \frac{T_f^2 z^2 Q}{4\mu^2}\text{cn}^2(\frac{T_fz}{\mu}\tau+\eta,\frac{Q}{2}) = \frac{T_f^2}{2\mu^2}\left(  -z\sqrt{\frac{Q}{2}}\text{cn}(u,m) +C_{\lambda_3} \right)^2,
                \label{eq:clam34}
\end{equation} 
which can only be satisfied if $C_{\lambda_3} = 0$.

From the transversality condition:
\begin{align}
    \lambda_3(1) & = -z\sqrt{\frac{Q}{2}}\text{cn}(\frac{T_fz}{\mu}+\eta,m)=0  
    \label{eq:lam_tran_con}
\end{align}
we have:
\begin{equation}
\tcbhighmath[drop fuzzy shadow]{\frac{T_fz}{\mu}+\eta  = (2n+1)\mathcal{K}(m).}
\label{eq:etaclf}
\end{equation}
where $\mathcal{K}(m)$ is the
complete elliptic integral of the first kind and $n$ is a natural number reflecting the periodicity of when the $cn$ function is zero.

Finally, the fourth costate $\lambda_4(\tau)$ (Equations~\ref{eq:lambda_4}-\ref{eq:cf_lam4}) is:
\begin{equation}
\begin{split}
\lambda_4(\tau) & = -\tau+\mu\tau+\frac{z^2}{2\mu}\frac{Q}{2}\tau+C_{\lambda_4} \label{eq:lam4cf}\\
\end{split}
\end{equation}
with the transversality condition requiring:
\begin{equation}
\begin{split}
\lambda_4(0) & = C_{\lambda_4} =0, \\
\lambda_4(1)& = -1+\mu+\frac{z^2}{2\mu}\frac{Q}{2}+C_{\lambda_4} = 0,
\end{split}
\end{equation}
we arrive at a closed form solution of $Q$:
\begin{equation}
\tcbhighmath[drop fuzzy shadow]{z^2=\frac{4\mu-4\mu^2}{Q}.}
\label{eq:Qclf}
\end{equation}

The cost function can now be simplified to:
\begin{equation}
    \begin{split}
    J & = \int _0^{1}T_f\Big((1-\mu)+\frac{\mu}{2}(v^2+\omega^2)\Big)d\tau = T_f-T_f\mu+\frac{T_fz^2Q}{4\mu}.
    \end{split}
\end{equation}
Substituting in $z^2$ into the cost function, it can be shown to reduce to 0. This implies that the determination of $Q$ and $T_f$ is a problem of  simultaneous solving the nonlinear Equations~\ref{eq:const1} and \ref{eq:const2} to satisfy the terminal constraints on the position of the WMR.
\begin{figure}[ht]     
\centering
    \includegraphics[width=0.55\textwidth]{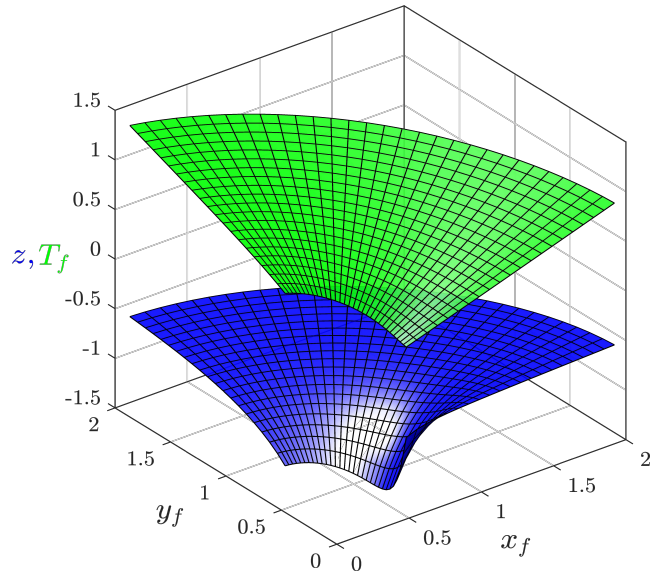}
    \caption{$z$ and $T_f$ displacement as a function of $x_f$ and $y_f$} 
    \label{fig:ztf_surf}
\end{figure}

Figure~\ref{fig:ztf_surf} illustrates the variation of the variables $z$ and $T_f$ as a function of the desired final $x$ and $y$ coordinates. The intersection of a vertical line centered at specific $x_f,y_f$ coordinate, with the two surfaces correspond to the optimal parameters $z$ and the maneuver time $T_f$. These two variables parametrize the optimal control and states.

 
\subsection{Second Formulation}

To design an optimal control which minimizes the weighted Energy-Time cost given by Equation~\ref{eq:costfn_tau}
for the system model described by Equations~\ref{eq:1WMB}-\ref{eq:3WMB},
we redefine the system in terms of the non-dimensional time $\tau = \frac{t}{T_f}$ as:
\begin{align}
\text{c}\theta\dot{x} +\text{s}\theta \dot{y} &= T_fv    \\
-\text{s}\theta\dot{x} +\text{c}\theta \dot{y} &= 0 \\
\dot {\theta} &= T_f\omega \\ 
\dot {T_f} &= 0 
\end{align}    
since $T_f$ is an unknown constant. $\text{c}\theta$ and $\text{s}\theta$ will be used as abbreviations for $\cos(\theta)$ and $\sin(\theta)$ respectively in the rest of this section.    

Using Lagrange multiplier to account for the kinematic constraints, results in the augmented cost function:
{\footnotesize \begin{align}
J_a & = \int_{0}^{1} \Big( T_f-T_f\mu + \frac{T_f\mu v^2}{2}+\frac{T_f\mu \omega^2}{2} +\zeta_1 (T_f v -\text{c}\theta\dot{x}-\text{s}\theta\dot{y})+ \zeta_2 (\text{s}\theta\dot{x}-\text{c}\theta\dot{y}) 
+ \zeta_3 (T_f\omega -\dot{\theta}) 
- \zeta_4 \dot{T_f} \Big) d\tau.
\end{align}}
The first variation of the augmented cost $J_a$ results in the necessary conditions for optimality:
\begin{align}
(\text{c}\theta\dot{x} +\text{s}\theta \dot{y}) & = T_f v \\
\text{s}\theta\dot{x}-\text{c}\theta\dot{y} & = 0 \\
\dot{\theta} & = T_f \omega\\
\dot{T_f} & = 0\\
v  &= -\frac{\zeta_1}{\mu} \label{eq:v_2}\\
\omega  &= -\frac{\zeta_3}{\mu} \label{eq:omg_2}\\
\dot{\zeta_1} &= -\frac{T_f\zeta_2\zeta_3}{\mu} \label{eq:zeta1}\\
\dot{\zeta_2} &= \frac{T_f\zeta_1\zeta_3}{\mu} \label{eq:zeta2}\\
\dot{\zeta_3} &=-\zeta_1(\text{s}\theta\dot{x}-\text{c}\theta\dot{y})-\zeta_2(\text{c}\theta\dot{x}+\text{s}\theta\dot{y}) 
 = -T_f\zeta_2 v  
 =  \frac{T_f\zeta_1\zeta_2}{\mu} \label{eq:zeta3}\\
\dot{\zeta_4} &= -(1-\mu +\frac{\mu}{2}v^2+\frac{\mu}{2}\omega^2+\zeta_1 v +\zeta_3 \omega ) \label{eq:zeta4} 
\end{align}
and the transversality conditions (TC):
\begin{align}
   x(0) &= 0, & y(0) &= 0, & \theta(0) &= 0,  \\
    x(1) &= x_f, & y(1) &= y_f, & &\\
    \zeta_3(1) &= 0, & \zeta_4(0) &= \zeta_4(1). & &
\end{align}
The motivation for presenting the two formulations will be clear in this section where interesting constraints on the evolving costates of the two formulations are derived.

Substitute Equations~\ref{eq:v_2} and \ref{eq:omg_2} into Equation~\ref{eq:zeta4} leads to:
\begin{equation}
    \dot{\zeta_4} = -1+\mu +\frac{\zeta_1^2}{2\mu}+\frac{\zeta_3^2}{2\mu}
    \label{eq:zeta_4dot}
\end{equation}
whose time derivative is:
\begin{equation}
    \ddot{\zeta_4} =  \dot{\zeta}_1\frac{\zeta_1}{\mu}+\dot{\zeta_3}\frac{\zeta_3}{\mu} = -\frac{T_f\zeta_2\zeta_3}{\mu}\frac{\zeta_1}{\mu}+\frac{T_f\zeta_1\zeta_2}{\mu}\frac{\zeta_3}{\mu} = 0,
\end{equation}
which requires $\dot{\zeta}_4$ be a constant. Since the transversality conditions requires $\zeta_4(0)=\zeta_4(1)$, this can only be satisfied when $\dot{\zeta}_4=0$. Equation~\ref{eq:zeta_4dot}, now is reduced to the constraint:
\begin{equation}
\tcbhighmath[drop fuzzy shadow]{\frac{1}{2}\zeta_3^2 + \frac{1}{2}\zeta_1^2 = \left( \sqrt{2\mu(1-\mu)} \right)^2 = K^2.}
\label{eq:cyl_con1}
\end{equation}
where the variable $K = \sqrt{2\mu(1-\mu)}$ will be used in the interest of brevity in the following development.

Pre-multiplying a vector comprised of Equations~\ref{eq:zeta1} and \ref{eq:zeta2} by $\begin{Bmatrix} \zeta_1 & \zeta_2 \end{Bmatrix}$, leads to:
\begin{eqnarray}
\begin{Bmatrix} \zeta_1 & \zeta_2 \end{Bmatrix} \begin{Bmatrix}\dot{\zeta_1} \\ \dot{\zeta_2} \end{Bmatrix} =  \begin{Bmatrix} \zeta_1 & \zeta_2 \end{Bmatrix} \begin{bmatrix} -\zeta_2 \\ \zeta_1
\end{bmatrix} \frac{T_f \zeta_3}{\mu} = 0
\end{eqnarray}
which can be rewritten as:
\begin{equation}
\zeta_1\dot{\zeta_1} +\zeta_2\dot{\zeta_2} = \frac{1}{2}\frac{d}{dt} \left(\zeta_1^2 + \zeta_2^2 \right)  = 0 
\end{equation}
which leads to:
\begin{equation}
\tcbhighmath[drop fuzzy shadow]{
\zeta_1^2 + \zeta_2^2  = \epsilon^2}
 \label{eq:cyl_con2}
\end{equation}
where $\epsilon$ is a constant. Equations~\ref{eq:cyl_con1} and~\ref{eq:cyl_con2} imply that the optimal costates $\zeta_1(t), \zeta_2(t)$ and $\zeta_3(t)$ lie on the intersection of two cylinders shown in Figure~\ref{fig:opt_cos_cyl}. Maclean and Biggs~\cite{maclean2013path} using optimal control theory for systems that are defined on Lie groups, illustrate the constraints provided by the Hamiltonian and the Casimir function generate intersecting cylinders as well. They consider a minimum energy optimal controller with a prescribed maneuver time unlike the cost function considered in this paper. 


Equations~\ref{eq:cyl_con1} and~\ref{eq:cyl_con2} 
can be used to derive the analytical expression for the costates.
The first costate equation~\ref{eq:zeta1} can be rewritten as: 
\begin{align}
\dot{\zeta_1} & = -\zeta_2\zeta_3\frac{T_f}{\mu} \nonumber\\
& = -\sqrt{\epsilon^2-\zeta_1^2}\sqrt{K^2-\zeta_1^2}\sqrt{\frac{T_f^2}{\mu^2}} \nonumber\\
& = -\sqrt{\frac{T_f^2}{\mu^2}\zeta_1^4-\frac{T_f^2 \epsilon^2 }{\mu^2}\zeta_1^2-\frac{T_f^2K^2}{\mu^2}\zeta_1^2+\frac{T_f^2\epsilon^2K^2}{\mu^2}}.  
\end{align}
The square of above expression leads to:
\begin{align}
\dot{\zeta_1}^2& = \frac{T_f^2}{\mu^2}\zeta_1^4-\frac{T_f^2 \epsilon^2 }{\mu^2}\zeta_1^2-\frac{T_f^2K^2}{\mu^2}\zeta_1^2+\frac{T_f^2\epsilon^2K^2}{\mu^2}\nonumber \\ 
& =\frac{T_f^2}{\mu^2}\zeta_1^4-\frac{T_f^2 \epsilon^2 }{\mu^2}(1+\frac{K^2}{\epsilon^2})\zeta_1^2+\frac{T_f^2\epsilon^2K^2}{\mu^2},  
\end{align}
which fits the form of the nonlinear ordinary differential equation whose solution is the Jaocbi elliptic $\text{sn}(.,m)$ function. The closed-form expression of $\zeta_1$ can now be represented as:
\begin{equation}
\tcbhighmath[drop fuzzy shadow]{
	\zeta_1(\tau) = K\text{sn}(\frac{T_f\epsilon}{\mu}\tau+\eta,\frac{K^2}{\epsilon^2})}
\label{eq:p2p_z1_closed}
\end{equation}
where $K$ and $\epsilon$ can be positive or negative. Using Equation~\ref{eq:cyl_con2} leads to the closed-form expression of $\zeta_2$ :
\begin{equation}
\tcbhighmath[drop fuzzy shadow]{
	\zeta_2(\tau) =  \epsilon \text{dn}(\frac{T_f\epsilon}{\mu}\tau+\eta,\frac{K^2}{\epsilon^2})}
\label{eq:p2p_z2_closed}
\end{equation}
where $dn(.,m)$ is the elliptic delta amplitude function. Similarly, using Equation~\ref{eq:cyl_con1}, we have:
\begin{equation}
\tcbhighmath[drop fuzzy shadow]{
	\zeta_3(\tau) =  K\text{cn}(\frac{T_f\epsilon}{\mu}\tau+\eta,\frac{K^2}{\epsilon^2}).}
\label{eq:p2p_z3_closed}
\end{equation}
The transversality condition requires $\zeta_3(1) = 0 $ which results in: 
\begin{align}
\eta = \mathcal{K}(m)-\frac{T_f\epsilon}{\mu}.
\label{eq:sec2_eta}
\end{align}
With closed from expressions for $\zeta_1$ and $\zeta_3$, the control variables can be solved as:
\begin{equation}
\tcbhighmath[drop fuzzy shadow]{
	v(\tau) = -\frac{K}{\mu}\text{sn}(\frac{T_f\epsilon}{\mu}\tau+\eta,\frac{K^2}{\epsilon^2})}
\label{eq:p2p_z_cf_v}
\end{equation}
\begin{equation}
\tcbhighmath[drop fuzzy shadow]{
	\omega(\tau) = -\frac{K}{\mu}\text{cn}(\frac{T_f\epsilon}{\mu}\tau+\eta,\frac{K^2}{\epsilon^2}).}
\label{eq:p2p_z_cf_w}
\end{equation}
Since $\dot{\theta}=T_f\omega$, replace $\omega$ with its closed-form which is stated in Equation~\ref{eq:p2p_z_cf_w} leading to: 
\begin{equation}
\tcbhighmath[drop fuzzy shadow]{
	\theta(\tau) =- \arccos(\text{dn} (u,m))+C_\theta}
\label{eq:p2p_z_cf_th}
\end{equation}
where $u=\frac{T_f\epsilon}{\mu}\tau+\eta$ and $m = \frac{K^2}{\epsilon^2}$  and the initial condition is given by:
\begin{align}
C_\theta = \arccos(\text{dn} (\eta,m)).
\end{align}
Using Equation~\ref{eq:p2p_z_cf_th}, the rate of change of position $x$ and $y$ (i.e $\dot{x}$ and $\dot{y}$) can be rewritten as :
\begin{align}
\dot{x}(\tau) & =\frac{T_fK}{\mu}\Biggr(\cos(C_\theta)\text{sn}(u,m)\text{dn}(u,m)+\sin(C_\theta)\sqrt{m}\text{sn}^2(u,m)  \Biggr) \\
\dot{y}(\tau) & = \frac{T_fK}{\mu}\Biggr(\sin(C_\theta)\text{sn}(u,m)\text{dn}(u,m)-\cos(C_\theta)\sqrt{m}\text{sn}^2(u,m)  \Biggr)
\end{align} 
which can be integrated and closed-form expression can be derived: 
\begin{equation}
\tcbhighmath[drop fuzzy shadow]{
	x(\tau) =\sin(C_\theta)(u-E(amp(u,m),m))-\frac{K \cos(C_\theta)}{\epsilon}\text{cn}(u,m)+C_x}
\label{eq:p2p_z_cf_x}
\end{equation}
\begin{equation}
\tcbhighmath[drop fuzzy shadow]{
	y(\tau) =- \cos(C_\theta)(u-E(amp(u,m),m))-\frac{K \sin(C_\theta)}{\epsilon}\text{cn}(u,m)+C_y}
\label{eq:p2p_z_cf_y}
\end{equation}
where $C_x$ and $C_y$ are the integration constant which can be defined by the initial conditions: 
\begin{align}
C_x = -\sin(C_\theta)(u-E(amp(u,m),m))+\frac{K \cos(C_\theta)}{\epsilon}\text{cn}(u,m)\\
C_y = \cos(C_\theta)(u-E(amp(u,m),m))+\frac{K \sin(C_\theta)}{\epsilon}\text{cn}(u,m)
\end{align}

To solve for any point-to-point Energy-Time optimal maneuver where the terminal point is given in polar form with the coordinates $(r,\alpha)$, the optimal control can be determined by solving two simultaneous nonlinear equations:
\begin{align}
r\cos(\alpha) &= \sin(C_\theta)(\mathcal{K}(m)-E(\frac{\pi}{2},m)) +C_x\\
r\sin(\alpha) &= -\cos(C_\theta)(\mathcal{K}(m)-E(\frac{\pi}{2},m))+C_y
\end{align}
for $\epsilon$ and $T_f$, where  
$u=\frac{T_f\epsilon}{\mu}\tau+\eta$ , $ m = \frac{K^2}{\epsilon^2}$ ,$\eta = \mathcal{K}(m)-\frac{T_f\epsilon}{\mu}$ and $K = \pm\sqrt{2\mu(1-\mu)}$.

\section{Results}

\begin{figure}
\centering
\subfloat[Optimal Trajectories]{
\includegraphics[width=55mm,height=51mm]{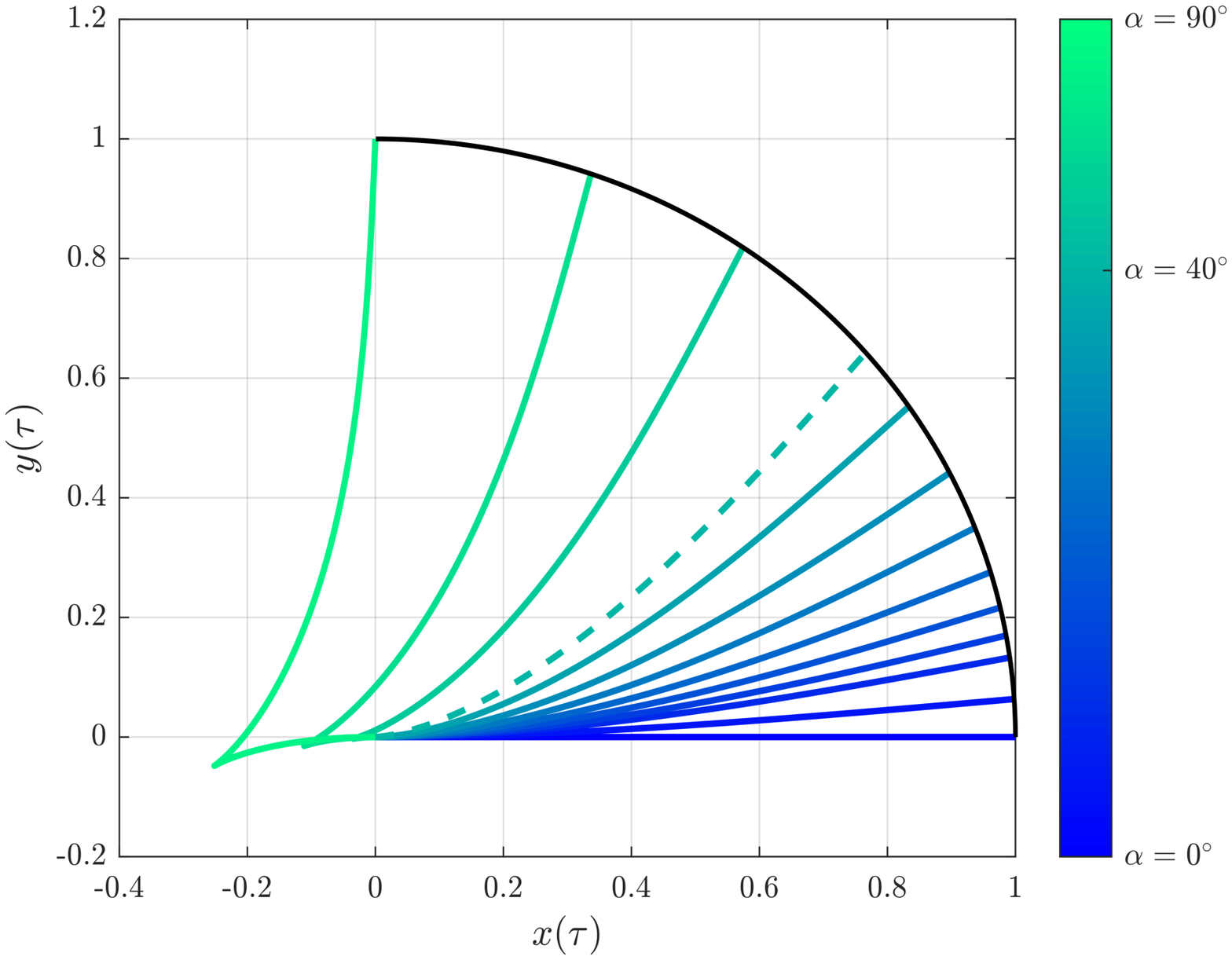}
\label{fig:energy_time_traj}}
\subfloat[Control Cylinder]{
  \includegraphics[width=55mm,height=55mm]{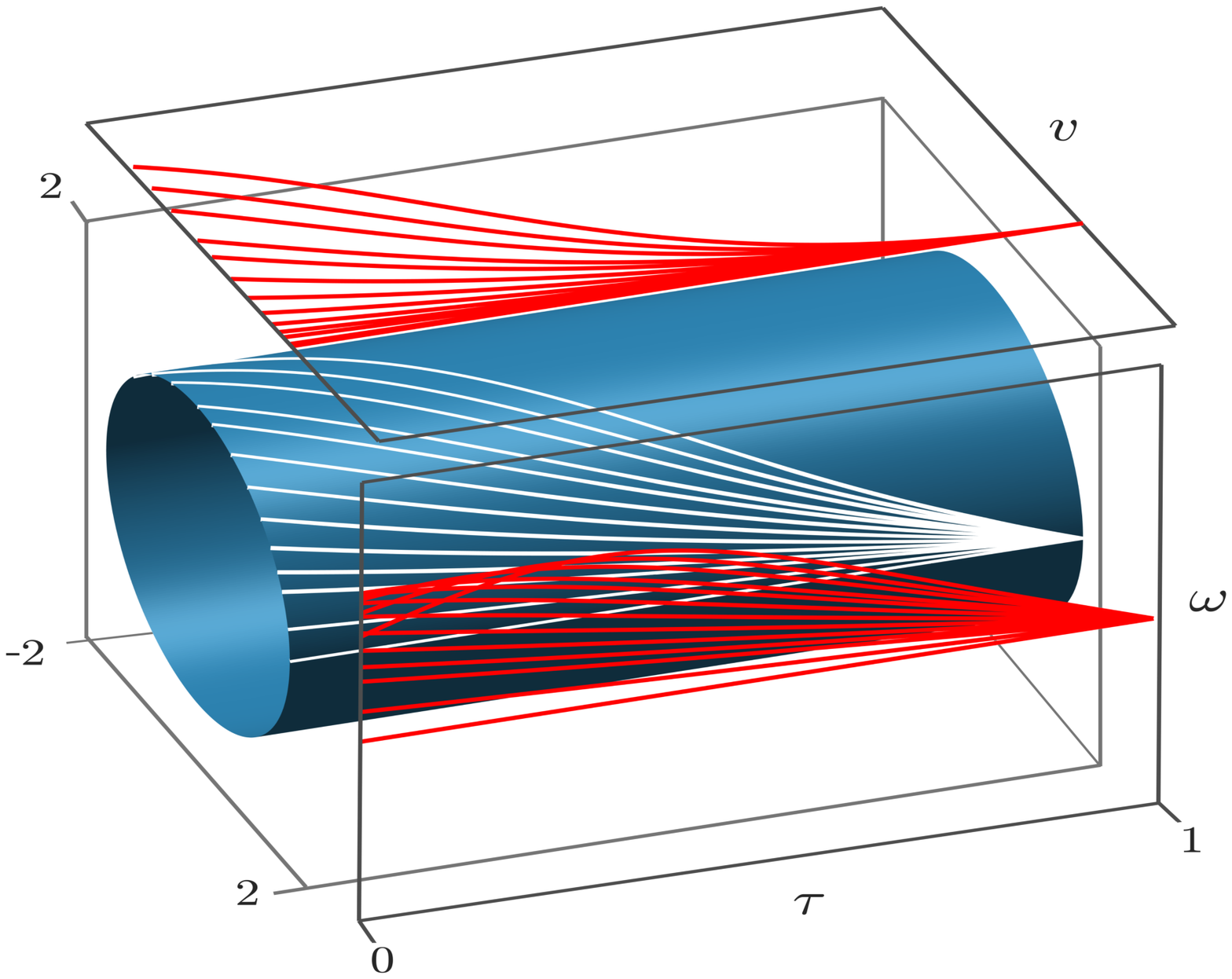}
\label{fig:cont_surf}}
\hspace{0mm}
\subfloat[Optimal control $v$]{
  \includegraphics[width=55mm,height=51mm]{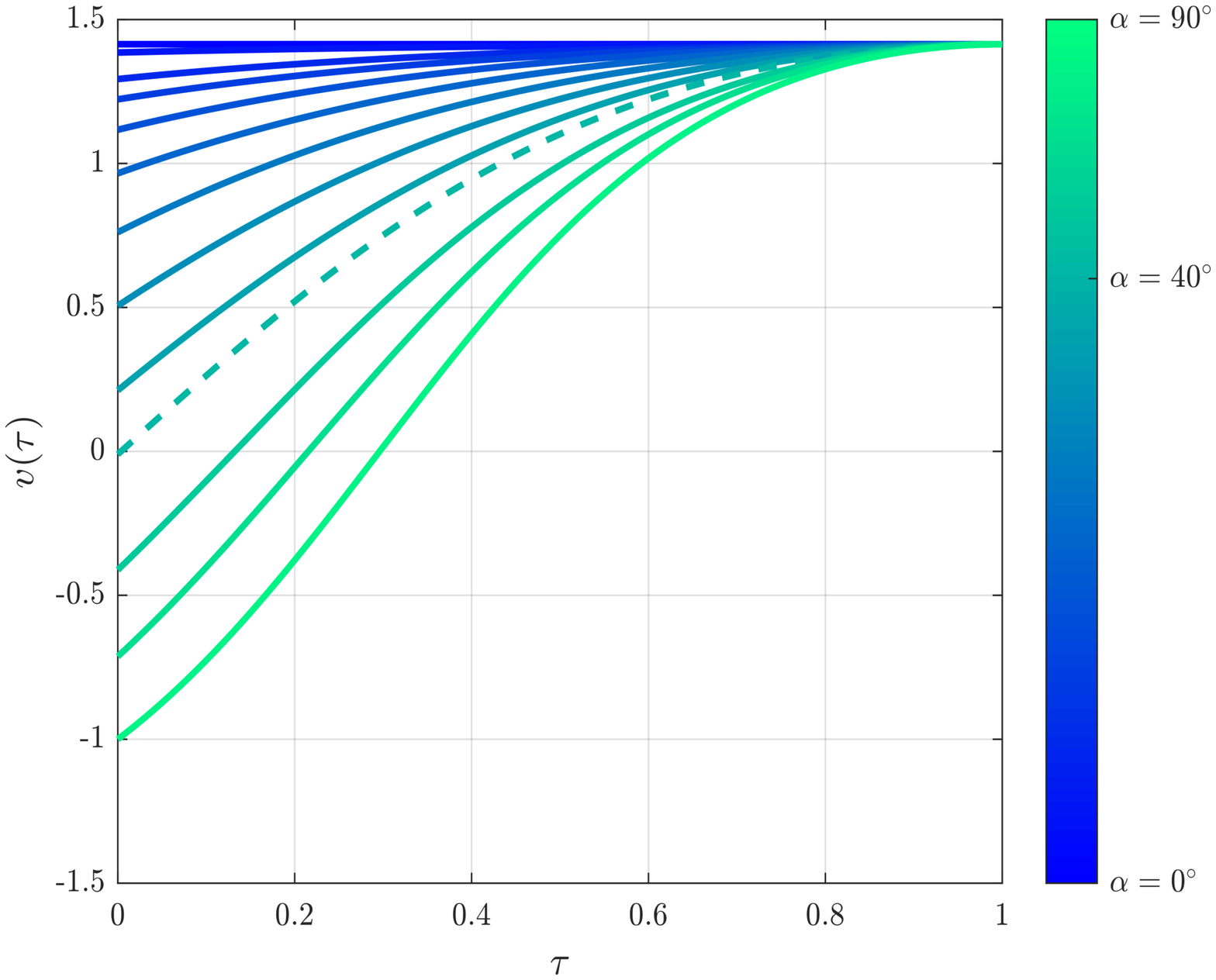}
\label{fig:energy_time_v}}
\subfloat[Optimal control $\omega$]{
  \includegraphics[width=55mm, height=52mm]{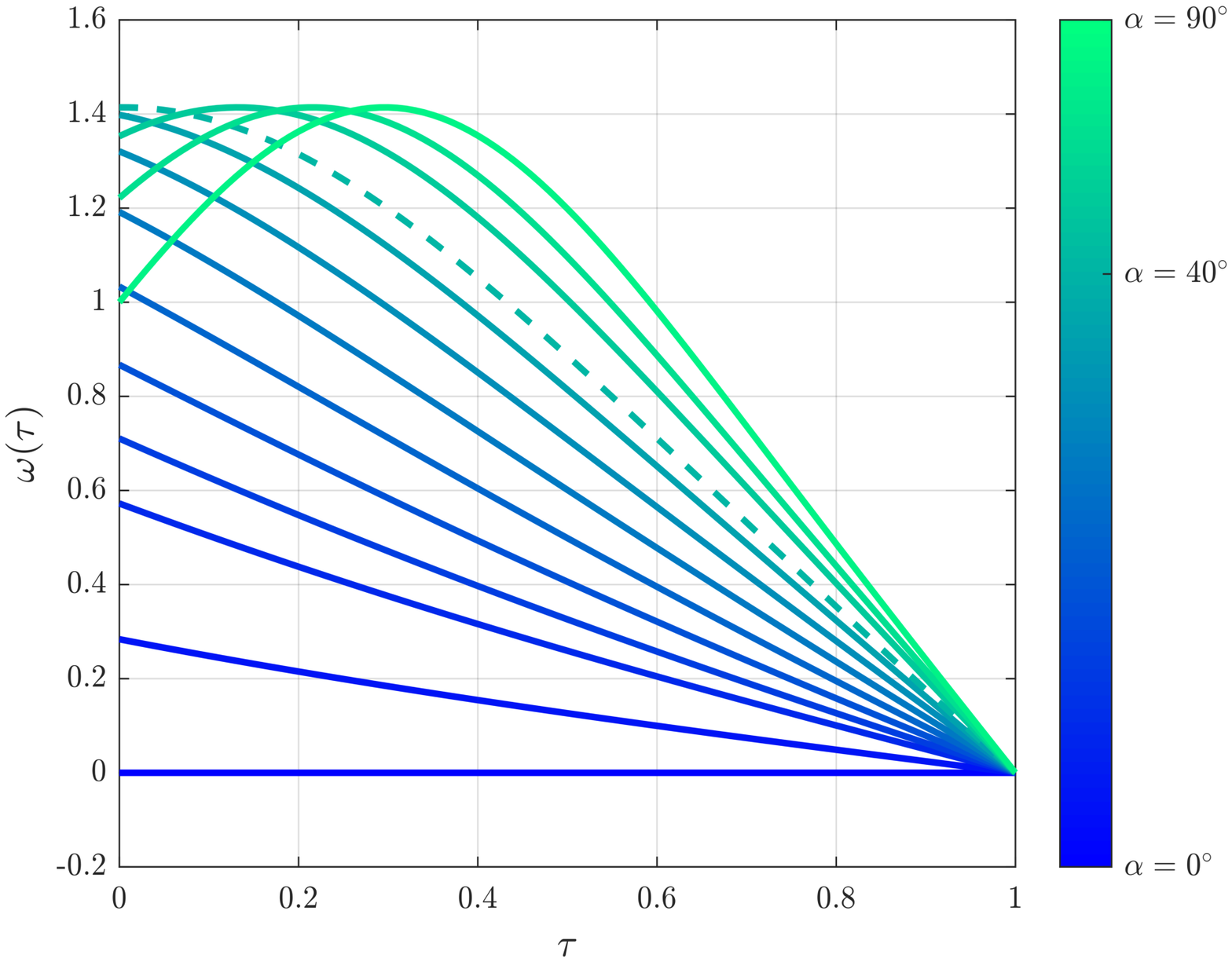}
\label{fig:energy_time_w}}
\caption{Energy-time optimal trajectories and control [R = 1 , $\mu =0.5$]}
\label{fig:energy_time_sol}
\end{figure}

To study the variation in the control as a function of terminal states, the terminal states are parameterized using polar coordinates. Fixing the radius of a circle on which the terminal states lie, the study considers the variation of the angle $\alpha$ of a vector connecting the origin to the terminal states.

\subsection{Optimal Control}

	Figure~\ref{fig:energy_time_traj} illustrates the optimal point-to-point motion trajectories all of which start at the origin and terminate on the arc of a circle of unit radius. Comprehensive numerical simulations were used to illustrate the computational benefit of the proposed approach. It was noted that for about  99\% of testing scenarios, the traditional shooting method did not converge to the optimal solution as opposed to a 100\% of convergence rate for the nonlinear programming approach. Furthermore, the average computational time required for convergence of the nonlinear programming approach was about 1.93 seconds while the shooting method took 4.18 seconds when the algorithm converged (detailed information can be found in Appendix B). Figures~\ref{fig:energy_time_v} and \ref{fig:energy_time_w} illustrate the time variation of the optimal linear and angular velocities of the WMR. Figure~\ref{fig:cont_surf} provides the combined plot of $v$ and $\omega$ as a function of time and is shown to lie on a cylinder. It should be noted that the optimal linear velocities for some trajectories start with a negative value, implying that the robot initially backs into the third quadrant before terminating in the first quadrant. 
It can also be noted from Figure~\ref{fig:energy_time_v} that at the terminal time of $\tau=1$, the optimal linear velocity for all trajectories is the same and from Figure~\ref{fig:energy_time_w} it can be noted that the optimal angular velocity is zero at the terminal time. Equation~\ref{eq:lam_tran_con} which corresponds to the transversality condition $\lambda_3(1)=0$ confirms the constraint that $\omega(\tau=1) = \frac{-\lambda_3}{\mu} = 0$. It was also shown in Equation~\ref{eq:Ham_constant} that the Hamiltonian is zero for all time resulting in the constraint:
\begin{equation}
  v^2(\tau) + \omega^2(\tau) =   \left( \frac{\sqrt{2\mu(1-\mu)}}{\mu} \right)^2
 \end{equation}
 which results in 
\begin{equation}
  v(1) = \left( \frac{\sqrt{2\mu(1-\mu)}}{\mu} \right).
 \end{equation} 

The closed form solution for the optimal angular velocity has been shown to be:
\begin{align}
 \omega &= -\frac{1}{\mu}\lambda_3 = \frac{z}{\mu}\sqrt{\frac{Q}{2}}\text{cn}(\frac{T_fz}{\mu}\tau +\eta ,\frac{Q}{2}).  
\end{align}
To determine when the angular velocity is stationary, we evaluate:

\begin{align}
\frac{d\omega}{d\tau} &= -\frac{T_fz^2}{\mu^2}\sqrt{\frac{Q}{2}}\text{sn}(\frac{T_fz}{\mu}\tau +\eta,\frac{Q}{2})\text{dn}(\frac{T_fz}{\mu}\tau +\eta,\frac{Q}{2}) =0  \nonumber \\ 
\Rightarrow \quad & \text{sn}(\frac{T_fz}{\mu}\tau +\eta,\frac{Q}{2}) = 0  \\
\therefore \tau & = -\eta\frac{\mu}{T_fz}
\end{align}

since $\text{sn}(0,m) = 0$.

Since the optimal linear velocity is:
\begin{align}
    v & = \frac{-z}{\mu}\sin(\theta+\phi) =\frac{-z}{\mu}\sin(\psi)  = \frac{-z}{\mu}\sin\bigg( \arcsin(\sqrt{\frac{Q}{2}}\text{sn}(\frac{T_fz}{\mu}\tau +\eta,\frac{Q}{2}) \bigg),
\end{align}
evaluating the linear velocity at $\tau = -\eta\frac{\alpha}{T_fz}$ we have:
\begin{align}
    v(-\eta\frac{\mu}{T_fz}) = 0
\end{align}
which indicates that when the linear velocity is zero, the angular velocity is stationary. This property can be used to precisely identify what terminal constraint corresponds to the transition of optimal trajectories which completely lie in the first quadrant and those that move into the third quadrant before moving into the first quadrant. This would require $\eta = 0$, which from Equation~\ref{eq:etaclf} leads to the constraint:
\begin{equation}
T_f z  = \mu (2n+1)\mathcal{K}(m).
\label{eq:tran_alpha}
\end{equation}
The dashed line in Figures~\ref{fig:energy_time_traj}, \ref{fig:energy_time_v} and~\ref{fig:energy_time_w} correspond to the solution when $\eta = 0$ which separates optimal trajectories which lie exclusively in the first quadrant from those that lie in the first and third quadrants.

Finally, the constraint that requires the norm of the optimal control to lie on a circle and the coupling between the linear and angular velocities can be better comprehended by observing the orthographic projection of the optimal control that lie on the constraint cylinder shown in Figure~\ref{fig:cont_surf}. The $\tau$ axis illustrates the evolution of the optimal control moving from left to right in the figure.
The $v-\tau$ and $\omega-\tau$ planes as shown in Figure~\ref{fig:cont_surf} reveal the reasons for the maximum magnitude of linear and angular velocities to be equal to the radius of the constraint cylinder. The colorbar indicates the range of the variable $\alpha$ which parameterizes the terminal states and ranges from $0^o$ to $90^o$ in Figures~\ref{fig:energy_time_sol}(a), (c) and (d).


\subsection{Optimal Costates}

The optimal costates for the first problem formulation presented in Section~\ref{sec:Form1} resulted in the equations:
\begin{eqnarray}
    \lambda_1(\tau) & = C_1 \\
    \lambda_2(\tau) & = C_2 \\
    \lambda_3(\tau) & =   -z\sqrt{\frac{Q}{2}}\text{cn}(\frac{T_f z }{\mu}\tau+\eta,m).
\end{eqnarray}
which implies that in the $\lambda_1-\lambda_2-\lambda_3$ space, the optimal costates are constrained to lie on a collinear line which is parallel to the $\lambda_3$ axis.

It was also shown that the optimal costates for the second problem formulation were constrained to lie on the intersection of two cylinders given by the equations:
\begin{eqnarray}
   \zeta_1^2 + \zeta_2^2  & = \epsilon^2 \\ 
   \zeta_1^2 + \zeta_3^2  & = 2\mu(1-\mu) 
\end{eqnarray}
which permits writing the parameterized costates as:
\begin{eqnarray}
   \zeta_1  & = & \sqrt{2\mu(1-\mu)} \sin(\delta) \\
   \zeta_2  & = & \sqrt{\epsilon^2-2\mu(1-\mu) \sin^2(\delta)} \\
   \zeta_3  & = & \sqrt{2\mu(1-\mu)} \cos(\delta) 
\end{eqnarray}
where $0 \le \delta \le 2\pi$. A unique scenario is when the two intersecting cylinder in the $\zeta_1-\zeta_2-\zeta_3$ space are of the same diameter, i.e., $\epsilon = \sqrt{2\mu(1-\mu)} = K$. This corresponds to the elliptic modulus reducing to unity and the elliptic integral of the first kind goes to $\infty$ resulting in $\eta=\infty$ from Equation~\ref{eq:sec2_eta}. 

For a modulus $m=1$, the Jacobi elliptic functions transition to hyperbolic functions, resulting in the optimal equations:
\begin{align}
    &\omega(\tau) = 0, & \theta(\tau) = 0, && v(\tau) = \frac{\sqrt{2\mu(1-\mu)}}{\mu}, &\\
    & x(\tau) = \frac{\sqrt{2\mu(1-\mu)}}{\mu} \tau, & y(\tau) = 0, && T_f = \frac{\mu}{\sqrt{2\mu(1-\mu)}}, & \\
    & \zeta_1(\tau) = \sqrt{2\mu(1-\mu)}, & \zeta_2(\tau) = 0, &&\zeta_3(\tau) = 0, & 
\end{align}
which corresponds to the terminal coordinate lying on a line which is an extension of the initial orientation of the WMR.

\begin{figure}
\centering
\subfloat[Optimal Trajectories $\alpha=0^o$]{
  \  \includegraphics[width=55mm,height=55mm]{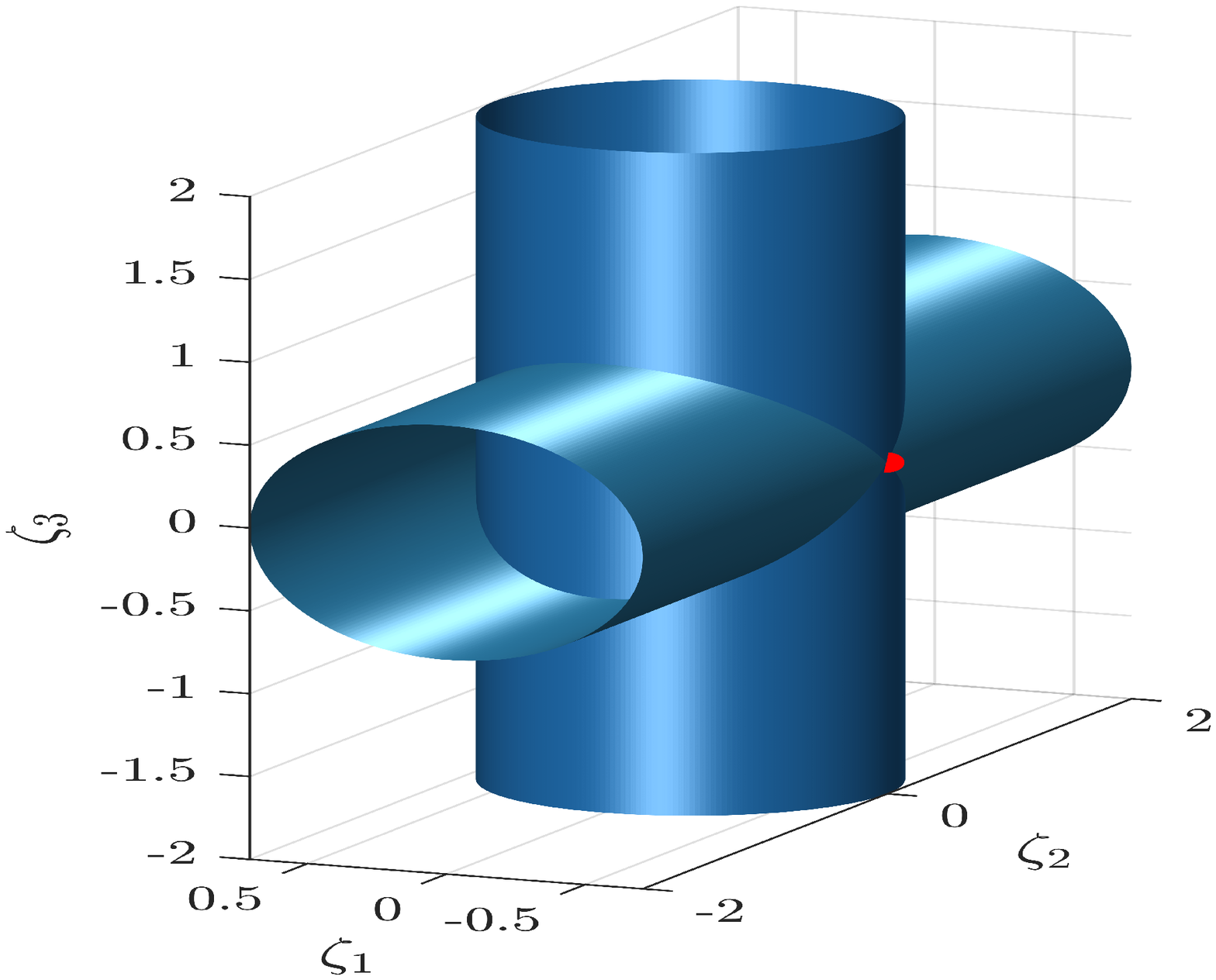}
\label{fig:energy_time_traj_CV0}}
\subfloat[Optimal Trajectories $\alpha=30^o$]{
  \includegraphics[width=55mm,height=55mm]{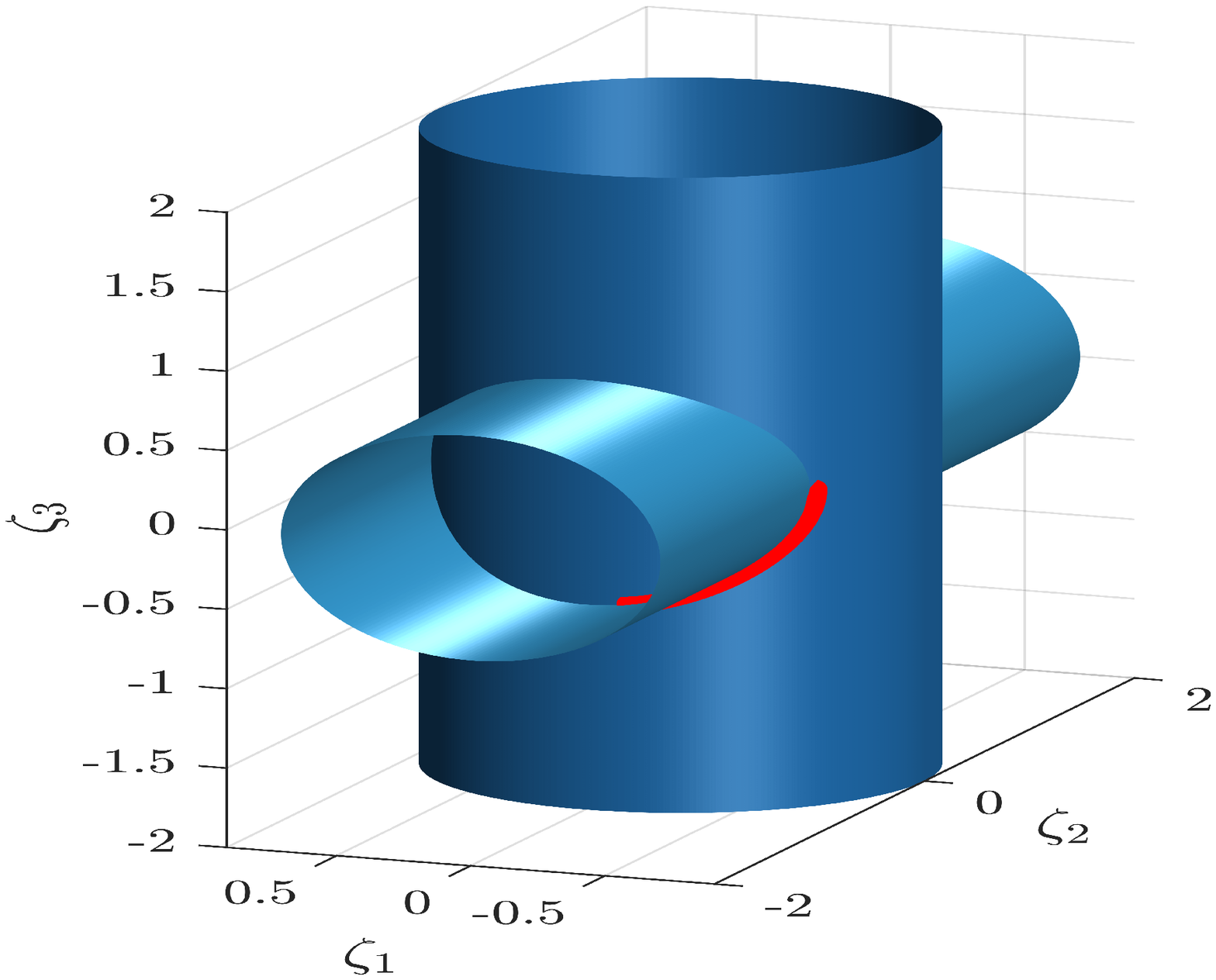}
\label{fig:energy_time_traj_CV30}}
\hspace{0mm}
\subfloat[Optimal Trajectories $\alpha=60^o$]{
  \includegraphics[width=55mm,height=55mm]{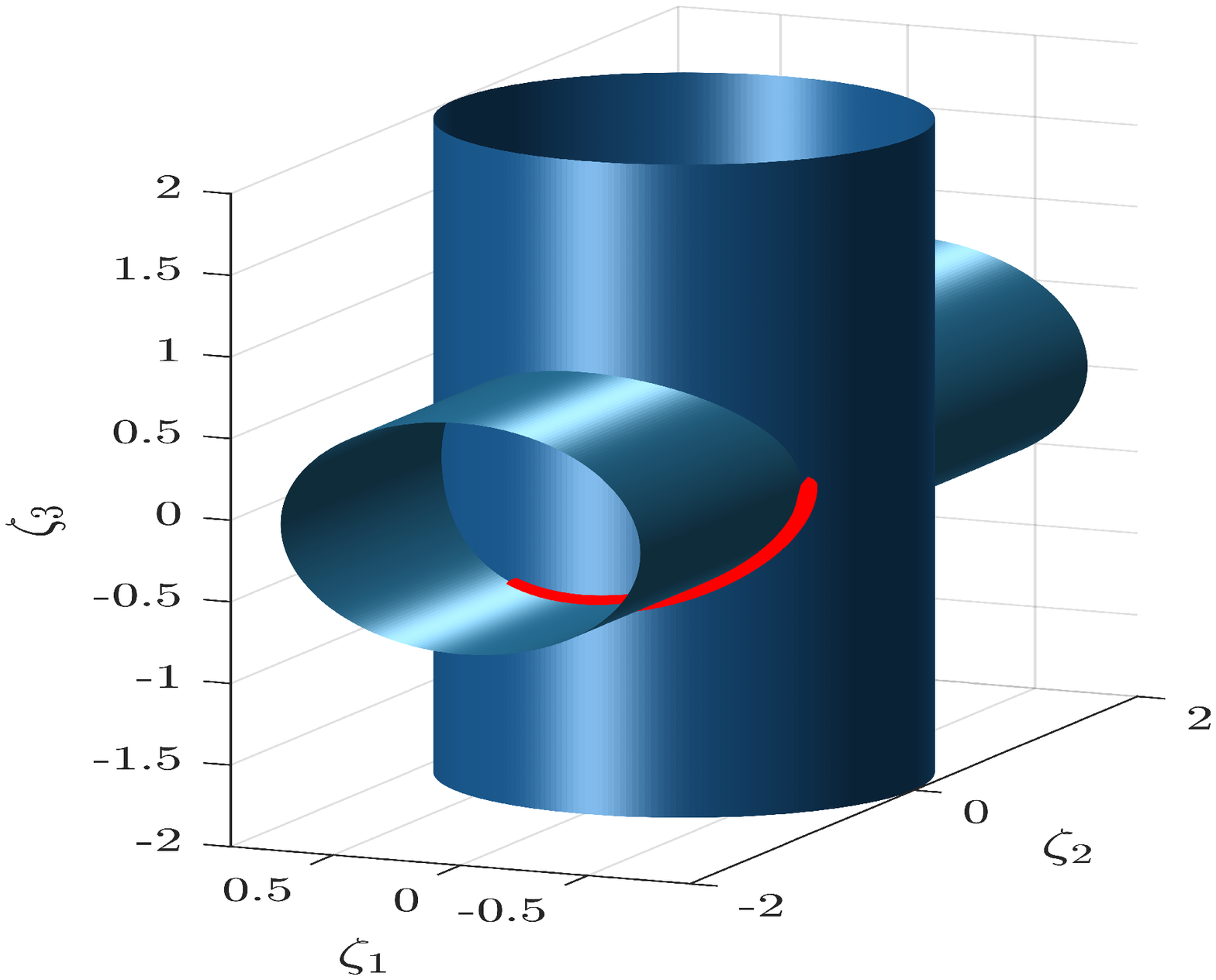}
\label{fig:energy_time_traj_CV60}}
\subfloat[Optimal Trajectories $\alpha=90^o$]{
  \includegraphics[width=55mm,height=55mm]{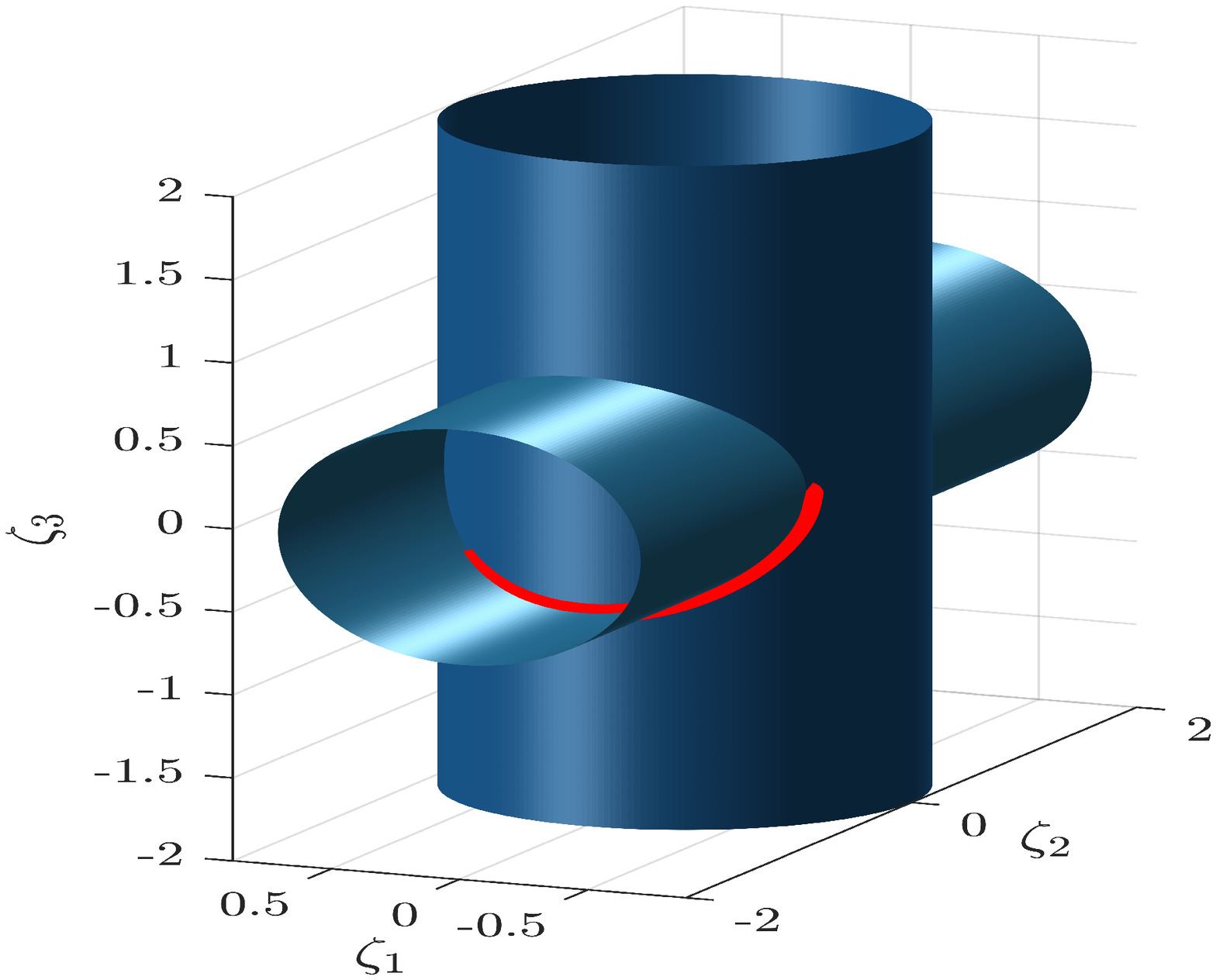}
\label{fig:energy_time_traj_CV90}}
\caption{Optimal Costates [R = 1 , $\mu =0.5$]}
\label{fig:opt_cos_cyl}
\end{figure}

Figure~\ref{fig:opt_cos_cyl} illustrates the evolution of the optimal costates for a representative set of maneuvers. The red markers correspond to the evolution of the costates $\zeta_1-\zeta_2-\zeta_3$ which are constrained to lie on the intersection of two cylinders. 
Optimal costate evolution are presented for maneuvers of $\alpha=0$, $\alpha=30^o$, $\alpha=60^o$, and $\alpha=90^o$. For the $\alpha=0$ maneuver, the costates for both formulations are time invariant and are coincident and located at the apex of the intersection of the two cylinders. For the rest of the maneuvers, it can be seen that the red line which is the line of intersection of the two cylinders is where the optimal costates $\zeta_1-\zeta_2-\zeta_3$ lie.

\section{Fixed Linear Velocity Robots}
The seminal work of Dubins~\cite{dubins1957curves} on shortest path for the unicycle model constrained the linear velocity to be constant and positive. The Reeds-Shepp work~\cite{reeds1990optimal} also constrained the linear velocity to be constant, but could be positive or negative. There is a large body of literature which considers the problem of path planning of wheeled mobile robots with constant linear velocity.  In this section, we consider the energy-time optimal control design where we constrain the linear velocity to be constant, but allow the optimal control solution to determine it. 
The cost function in the normalized $\tau = \frac{t}{T_f}$ space is: 
    \begin{align}\label{eq:vcont_cost}
       J=  \int _0^{1}T_f\left((1-\mu)+\frac{\mu}{2}(v_c^2+\omega^2)\right)d\tau \qquad  
    \end{align}
where $\mu$ is a weighting factor, bound between 0 to 1. The variable $v_c$ is the constant linear velocity input and $\omega$ is the angular velocity input. The mobile robot is constrained by the equations:
\begin{align}
\text{c}\theta\dot{x} +\text{s}\theta \dot{y} &= T_f v_c \label{eq:C_velEq1}\\
-\text{s}\theta\dot{x} +\text{c}\theta \dot{y} &= 0 \\
\dot {\theta} &= T_f\omega \label{eq:thetadot1} \\ 
\dot {T_f} &= 0 \\
\dot {v_c} & = 0 \label{eq:vcdot}
\end{align}
where Equation~\ref{eq:vcdot} enforces the constant linear velocity constraint. The initial and terminal state constraints are:

\begin{align}
    \begin{bmatrix}
x(0)\\ y(0) \\  \theta(0) \\
\end{bmatrix} = 
    \begin{bmatrix}
0\\ 0 \\  0 \\
\end{bmatrix} ,
\begin{bmatrix}
x(1)\\ y(1) \\  \theta(1) \\
\end{bmatrix} = 
    \begin{bmatrix}
r\cos(\alpha)\\r\sin(\alpha) \\  free \\
\end{bmatrix} ,
\begin{bmatrix}
T_f(0)\\ v_c(0)  
\end{bmatrix} = 
\begin{bmatrix}
T_f(1) \\ v_c(1) 
\end{bmatrix}.
\end{align}

The first variation of the augmented cost function:
\begin{align}
J_a & = \int_{0}^{1} \left( T_f-T_f\mu + \frac{T_f\mu v_c^2}{2}+\frac{T_f\mu \omega^2}{2}  
+\zeta_1 (T_f v_c -c\theta\dot{x}-s\theta\dot{y}) \right. \nonumber  \\ 
& \qquad \qquad \qquad + \zeta_2 (s\theta\dot{x}-c\theta\dot{y}) 
 \left. + \zeta_3 (T_f\omega -\dot{\theta}) 
+ \zeta_4 (-\dot{T_f}) 
- \zeta_5 \dot{v_c} \right) d\tau 
\end{align}
results in the equations: 
\begin{align}
w  &= -\frac{\zeta_3}{\mu} \label{eq:omg_sol} \\ 
(\dot{\zeta_1}\text{c}\theta-\zeta_1\text{s}\theta\dot{\theta})&=(\dot{\zeta_2}\text{s}\theta+\zeta_2\text{c}\theta\dot{\theta}) \label{eq:cos1}\\
(\dot{\zeta_1}\text{s}\theta+\zeta_1\text{c}\theta\dot{\theta})&=-(\dot{\zeta_2}\text{c}\theta-\zeta_2\text{s}\theta\dot{\theta}) \label{eq:cos2}\\
\dot{\zeta_3} &=-\zeta_1(\text{s}\theta\dot{x}-\text{c}\theta\dot{y})-\zeta_2(\text{c}\theta\dot{x}+\text{s}\theta\dot{y}) \label{eq:cos3}\\
\dot{\zeta_4} &= -(1-\mu +\frac{\mu}{2}v_c^2+\frac{\mu}{2}\omega^2+\zeta_1 v_c +\zeta_3 \omega  ) \label{eq:cos4} \\
\dot{\zeta_5} & =-T_f\mu v_c-T_f\zeta_1 \label{eq:cos5}
\end{align}
which in conjunction with the state equations (\ref{eq:C_velEq1}-\ref{eq:vcdot}) form the necessary conditions for optimality.
The associated transversality conditions are:

    \begin{align}
    \begin{bmatrix}
x(0)\\ y(0) \\  \theta(0) \\
\end{bmatrix} = 
    \begin{bmatrix}
0\\ 0 \\  0 \\
\end{bmatrix},
\begin{bmatrix}
x(1)\\ y(1) \\  \zeta_3(1) \\
\end{bmatrix} = 
    \begin{bmatrix}
r\cos(\alpha)\\r\sin(\alpha) \\  0\\
\end{bmatrix}, 
\begin{bmatrix}
\zeta_4(0)\\ \zeta_5(0)  
\end{bmatrix} = 
\begin{bmatrix}
\zeta_4(1) \\ \zeta_5(1) 
\end{bmatrix}.
\end{align}

Equations~\ref{eq:cos1} and \ref{eq:cos2} can be rewritten as:
\begin{eqnarray}
\begin{bmatrix} \text{c}\theta & -\text{s}\theta \\ \text{s}\theta & \text{c}\theta
\end{bmatrix} \begin{bmatrix}\dot{\zeta_1} \\ \dot{\zeta_2} \end{bmatrix} = \begin{bmatrix} \zeta_1 \text{s}\theta + \zeta_2 \text{c}\theta \\ \zeta_2 \text{s}\theta - \zeta_1 \text{c}\theta
\end{bmatrix} \dot{\theta} 
\end{eqnarray}
which leads to:
\begin{eqnarray}
 \begin{bmatrix}\dot{\zeta_1} \\ \dot{\zeta_2} \end{bmatrix} = \begin{bmatrix} \text{c}\theta & \text{s}\theta \\ -\text{s}\theta & \text{c}\theta
\end{bmatrix} \begin{bmatrix} \zeta_1 \text{s}\theta + \zeta_2 \text{c}\theta \\ \zeta_2 \text{s}\theta - \zeta_1 \text{c}\theta
\end{bmatrix} \dot{\theta} = \begin{bmatrix} \zeta_2 \\ - \zeta_1
\end{bmatrix} \dot{\theta} = \begin{bmatrix} \zeta_2 \\ - \zeta_1
\end{bmatrix} T_f \omega = \begin{bmatrix} -\zeta_2 \\ \zeta_1
\end{bmatrix} \frac{T_f \zeta_3}{\mu} \label{eq:cobilin}
\end{eqnarray}
Pre-multiply both sides by $\begin{bmatrix} \zeta_1 & \zeta_2 \end{bmatrix}$, we have:
\begin{eqnarray}
\begin{bmatrix} \zeta_1 & \zeta_2 \end{bmatrix} \begin{bmatrix}\dot{\zeta_1} \\ \dot{\zeta_2} \end{bmatrix} =  \begin{bmatrix} \zeta_1 & \zeta_2 \end{bmatrix} \begin{bmatrix} -\zeta_2 \\ \zeta_1
\end{bmatrix} \frac{T_f \zeta_3}{\mu} = 0
\end{eqnarray}
which can be rewritten as:
\begin{eqnarray}
\zeta_1\dot{\zeta_1} +\zeta_2\dot{\zeta_2} = \frac{d}{dt} \left(\zeta_1^2 + \zeta_2^2 \right)  = 0 \\
\tcbhighmath[drop fuzzy shadow]{\zeta_1^2 + \zeta_2^2 = K_c^2.} \label{eq:cylcon}
\end{eqnarray}

\begin{figure}
 \centering
 \subfloat[Optimal Trajectories]{
 \includegraphics[width=55mm,height=55mm]{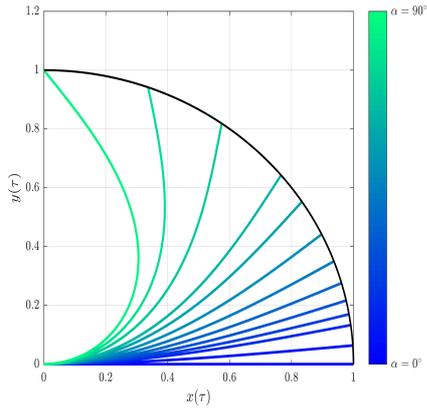}
 \label{fig:dubin_traj}}
 \subfloat[Costate constraints]{
  \includegraphics[width=55mm,height=55mm]{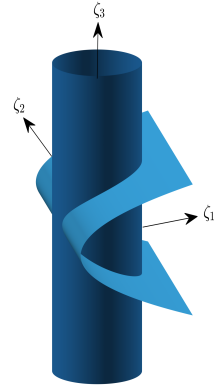}
    \label{fig:cylpar}} \\
 \subfloat[Optimal control $v_c$]{
 \includegraphics[width=55mm,height=55mm]{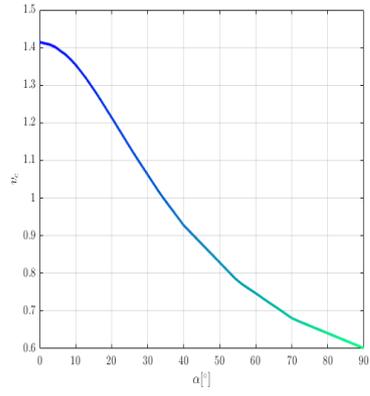}
 \label{fig:dubin_vc}}
 \subfloat[Optimal control $\omega$]{
   \includegraphics[width=55mm,height=55mm]{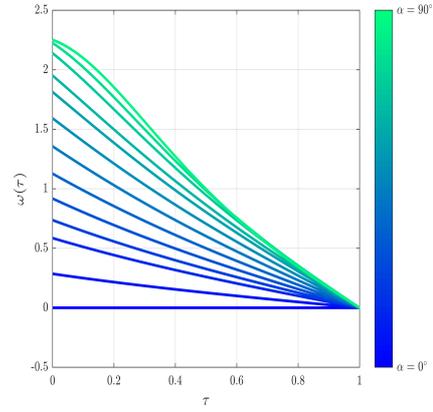}
 \label{fig:dubin_w}}
 \caption{Energy-time optimal trajectories and control for Fixed Linear velocity [R = 1 , $\mu =0.5$]}
 \label{fig:dubin_sol}
\end{figure}

Equation~\ref{eq:cos3} can be rewritten as:
\begin{equation}
\dot{\zeta_3} =-\zeta_2(\text{c}\theta\dot{x}+\text{s}\theta\dot{y}) = -\zeta_2 T_f v_c    
\label{eq:simpeqs}
\end{equation}
multiply both sides by $\zeta_3$, leading to:
\begin{equation}
\zeta_3 \dot{\zeta_3} =-\zeta_3 \zeta_2 T_f v_c    
\end{equation}
using Equation~\ref{eq:cobilin}, we now have:
\begin{equation}
\zeta_3 \dot{\zeta_3} = \frac{1}{2} \frac{d}{dt} \left( \zeta_3^2 \right) =\dot{\zeta}_1 \mu v_c    
\end{equation}
which can be simplified to:
\begin{equation}
\tcbhighmath[drop fuzzy shadow]{\frac{1}{2} \zeta_3^2 =\zeta_1 \mu v_c   +C_c} \label{eq:parcon}
\end{equation}
which is a parabolic equation. Equations~\ref{eq:cylcon} and \ref{eq:parcon}  constrain the optimal co-states $\zeta_1(t), \zeta_2(t)$ and $\zeta_3(t)$ to lie on the intersection of a cylinder and an extruded parabola as shown in Figure~\ref{fig:cylpar}. Justh and Krishnaprasad~\cite{justh2015optimality} arrived at the same constraint for the costates when optimizing the energy cost for a mobile robot with unit linear velocity using the Hamiltonian and Casimir function.



Closed form solution for the optimal velocity can be determined by examining Equation~\ref{eq:cos4} which can be rewritten using Equation~\ref{eq:omg_sol} and \ref{eq:parcon} as:
\begin{align}
    \dot{\zeta_4} &= -(1-\mu +\frac{\mu}{2}v_c^2-\frac{1}{2\mu}\zeta_3^2+\zeta_1 v_c   ) = -(1-\mu +\frac{\mu}{2}v_c^2-\frac{C_c}{\mu} -\cancelto{0}{\zeta_1 v_c}+\cancelto{0}{\zeta_1 v_c}   ).
\end{align}
Since the transversality constraint requires $\zeta_4(0) = \zeta_4(1)$ and since $\dot{\zeta_4}$ is a constant, this requires $\dot{\zeta_4}=0$, which leads to the equation:
\begin{equation}
\mu(1-\mu) + \frac{\mu^2}{2}v_c^2   = C_c, \label{eq:cc_2}     
\end{equation}
which leads to a closed-form expression for the constant linear velocity in terms of $\mu$ and $C_c$: 
\begin{align}
    v_c = \pm\frac{\sqrt{2\mu^2-2\mu+2C_c}}{\mu}.
\end{align}
Since we are considering motion in the first quadrant of WMR whose initial orientation is coincident with the $x$-axis, we only consider positive forward velocity.
Using the costates constraints from Equation~\ref{eq:cylcon}
and \ref{eq:parcon}, the differential equation for the first costate $\zeta_1$ can be rewritten as:
\begin{align}
\dot{\zeta_1} &= -\zeta_2\zeta_3\frac{T_f}{\mu} \nonumber\\
& = -\sqrt{K_c^2-\zeta_1^2}\sqrt{2\zeta_1\mu v_c +2C_c}\sqrt{\frac{T_f^2}{\mu^2}} \nonumber\\
& = -\sqrt{\frac{-2v_cT_f^2}{\mu}\zeta_1^3-\frac{2C_c T_f^2}{\mu^2}\zeta_1^2+\frac{2K_c^2 v_cT_f^2}{\mu}\zeta_1+\frac{2K_c^2 C_cT_f^2}{\mu^2}} \nonumber \\
\longrightarrow \dot{\zeta_1}^2 & = \frac{-2v_cT_f^2}{\mu}\zeta_1^3-\frac{2C_c T_f^2}{\mu^2}\zeta_1^2+\frac{2K_c^2 v_cT_f^2}{\mu}\zeta_1+\frac{2K_c^2 C_cT_f^2}{\mu^2} \nonumber\\
& =  \frac{-2v_cT_f^2}{\mu} \left( \zeta_1 +\frac{C_c}{v_c\mu}\right) \left( \zeta_1 +{K_c}\right) \left( \zeta_1 -{K_c}\right). \label{eq:eq:zeta1dot_2}
\end{align}
The transversality condition $\zeta_3(1) =0$, results in Equation~\ref{eq:parcon} reducing to:
\begin{equation}
    \zeta_1(1) = -\frac{C_c}{\mu v_c}
\end{equation}
which substituted into Equation~\ref{eq:cylcon}
 results in the equation:
\begin{equation}
    \zeta_2^2(1) = K_c^2-\left(\frac{C_c}{v_c \mu}\right)^2,
\end{equation}
which requires

\begin{equation}
    {K_c} \ge \frac{C_c}{v_c \mu},
    \label{eq:root_ineq}
\end{equation}
which implies the roots of the polynomial on the right hand side of  Equation~\ref{eq:eq:zeta1dot_2} are:
\begin{equation}
   \left( \alpha_1 = -{K_c} \right) \le \left( \alpha_2 = -\frac{C_c}{v_c \mu} \right)  \le \left( \alpha_3 = {K_c} \right) 
\end{equation}
where $K_c$, $C_c$, $v_c$ and $\mu$ are all positive numbers.
Define a variable $\Phi$ and a linear transformation:
\begin{equation}
   \Phi^2 = \frac{\zeta_1 - \alpha_3}{\alpha_2-\alpha_3} = \frac{\zeta_1 - {K_c}}{-\frac{C_c}{v_c \mu}-{K_c}} \label{eq:tranform}
\end{equation}
leading to Equation~\ref{eq:eq:zeta1dot_2} being rewritten as:
\begin{equation}
   \left(\frac{d\Phi}{d\tau}\right)^2 = \Lambda^2 \left(1-\Phi^2 \right)  \left( 1 -m \Phi^2\right). \label{eq:psidot}
\end{equation}
The resulting solution of Equation~\ref{eq:psidot} is: 
\begin{align}
    \Phi = \text{sn}(\Lambda\tau+\eta,m)
\end{align}
where 
\begin{align}
      m =\frac{\alpha_3-\alpha_2}{\alpha_3-\alpha_1} = \frac{{K_v} v_c \mu+C_c}{2{K_c} v_c \mu}\quad \text{and} \: \: \: \: \Lambda^2 =\frac{v_cT_f^2}{\mu}{K_c}. \label{eq:m_Lambda}
\end{align}
We now can use the transformation which was defined in Equation \ref{eq:tranform} to map back to $\zeta_1$:
\begin{equation}
\tcbhighmath[drop fuzzy shadow]{\zeta_1(\tau) = {K_c}-{K_c}\text{sn}^2(\Lambda\tau+\eta,m)-\frac{C_c}{v_c\mu}\text{sn}^2(\Lambda\tau+\eta,m)}\label{eq:zeta1_closed}
\end{equation}
 
Substituting Equation~\ref{eq:zeta1_closed} into \ref{eq:cylcon}, we have: 

\begin{equation}
\tcbhighmath[drop fuzzy shadow]{
        \zeta_2(\tau) =\sqrt{2K_c^2+\frac{2C_cK_c}{v_c\mu}}\text{sn}(\Lambda\tau+\eta,m)\text{dn}(\Lambda\tau+\eta,m).} \label{eq:zeta2_closed}
\end{equation}

Likewise, closed-form expression for $\zeta_3$ can be obtained by using Equation~\ref{eq:zeta1_closed} and~\ref{eq:parcon}:
\begin{equation}
\tcbhighmath[drop fuzzy shadow]{
\zeta_3(\tau) = \pm\sqrt{2\mu v_c {K_c}+2C_c} \,\text{cn}(\Lambda\tau+\eta,m).}
\label{eq:zeta3_closed}
\end{equation}
Since $\zeta_3(1) = 0$, it results in $\eta$ to be : 
\begin{align}
    \eta = \mathcal{K}(m) - \Lambda.
 \end{align}

We know the optimal angular velocity command $\omega = -\frac{\zeta_3}{\mu}$, such that control input $\omega$ can be derived in closed-form: 
\begin{equation}
\tcbhighmath[drop fuzzy shadow]{
\omega(\tau) = \mp\frac{\sqrt{2\mu v_c {K_c}+2C_c}}{\mu} \,\text{cn}(\Lambda\tau+\eta,m).} \label{eq:zeta3_closed2}
\end{equation}

Since we defined the terminal positions which lie on a circle in the first quadrant, such that $\omega$ must be positive (i.e. counter-clockwise), consequently the sign of $\zeta_3$ must be negative.

Replacing $\zeta_1$ by its closed-form expression in Equation~\ref{eq:cos5}: 
\begin{align}
\dot{\zeta_5}& = -T_f\mu v_c-T_f{K_c}+T_f{K_c}\text{sn}^2(u,m)+\frac{C_cT_f}{v_c\mu}\text{sn}^2(u,m)
\end{align}
which can be integrated resulting in the equation: 
\begin{equation}
\tcbhighmath[drop fuzzy shadow]{
    \zeta_5(\tau) = -T_f\mu v_c\tau-T_f{K_c}\tau+\frac{2{K_c}T_f}{\Lambda}(u-E(amp(u,m),m))+C_{\zeta_5}}\label{eq:zeta5_closed}
\end{equation}
where $ u = \Lambda\tau+\eta$ and $C_{\zeta_5}$ is a constant of integration. 

Equation~\ref{eq:zeta3_closed2} in conjunction with ~\ref{eq:thetadot1} results in a closed form solution (details provided in the  C, Equation \ref{eq_apx:dubin_thdot}-\ref{eq:dubin_cf_theta}):
\begin{equation}
\tcbhighmath[drop fuzzy shadow]{
	\theta(\tau) = 2\arccos(\text{dn}(\Lambda\tau+\eta,m))+C_\theta}\label{eq:theta_closed}
\end{equation}
with the initial condition 
\begin{align}
C_\theta=-2\arccos(\text{dn}(\eta,m)),
\end{align}
which leads to the solutions for $x(\tau)$ and   $y(\tau)$ (Equations \ref{eq:dubin_xdot}-\ref{eq:y_closed}):
\begin{equation}
\tcbhighmath[drop fuzzy shadow]{
	x(\tau) = T_f v_c \Biggr(\frac{2\cos(C_\theta)}{\Lambda}E(amp(u,m),m)-\cos(C_\theta)\tau+\frac{2\sqrt{m}\sin(
		C_\theta)}{\Lambda}\text{cn}(u,m) \Biggr)+C_x\label{eq:x_closed}}
\end{equation}
with the initial condition 
\begin{align}
C_x = -T_f v_c \Biggr(\frac{2\cos(C_\theta)}{\Lambda}E(amp(\eta,m),m)+\frac{2\sqrt{m}sin(
	C_\theta)}{\Lambda}\text{cn}(\eta,m) \Biggr)
\end{align}
and
\begin{equation}
\tcbhighmath[drop fuzzy shadow]{
	y(\tau) = T_f v_c \Biggr(\frac{2\sin(C_\theta)}{\Lambda}E(amp(u,m),m)-\sin(C_\theta)\tau-\frac{2\sqrt{m}\cos(C_\theta)}{\Lambda}\text{cn}(u,m) \Biggr)+C_y\label{eq:y_closed}}
\end{equation}
with the initial condition 
\begin{align}
C_y = -T_f v_c \Biggr(\frac{2\sin(C_\theta)}{\Lambda}E(amp(\eta,m),m)-\frac{2\sqrt{m}\cos(
	C_\theta)}{\Lambda}\text{cn}(\eta,m) \Biggr).
\end{align}

Solving for the optimal control now reduces to solving three simultaneous nonlinear equations in three unknown parameters which are $m$, $T_f$ and $v_c$:
\begin{align}
r\cos(\alpha) &= T_f v_c \Biggr(\frac{2\cos(C_\theta)}{\Lambda}E(\frac{\pi}{2},m)-\cos(C_\theta) \Biggr)+C_x\\
r\sin(\alpha) &= T_f v_c \Biggr(\frac{2\sin(C_\theta)}{\Lambda}E(\frac{\pi}{2},m)-\sin(C_\theta) \Biggr)+C_y\\
\frac{2{K_c}T_f}{\Lambda}(\eta-E(amp&(\eta,m),m))= -T_f\mu v_c-T_f{K_c}-\frac{2{K_c}T_f}{\Lambda}E(\frac{\pi}{2},m),
\end{align}
where $\eta = \mathcal{K}(m)-\Lambda$ and $K_c = \frac{C_c}{v_c\mu(2m-1)}$.

Figure~\ref{fig:dubin_sol} (a), (c) and (d) presents a set of optimal trajectories, linear velocities and time evolution of the optimal angular velocity of the Dubins' like WMR model. These solutions corresponds to terminal position of the robot constrained to a circle of radius unity and with a weighting parameter $\mu=0.5$. The colorbar associated with Figure~\ref{fig:dubin_sol} (a) and (d) permits associating specific graphs with the associated $\alpha$ which parameterizes the terminal state of the trajectory.

\begin{figure}
\centering
\subfloat[Optimal Trajectories $\alpha=0^o$]{
  \includegraphics[width=55mm,height=55mm]{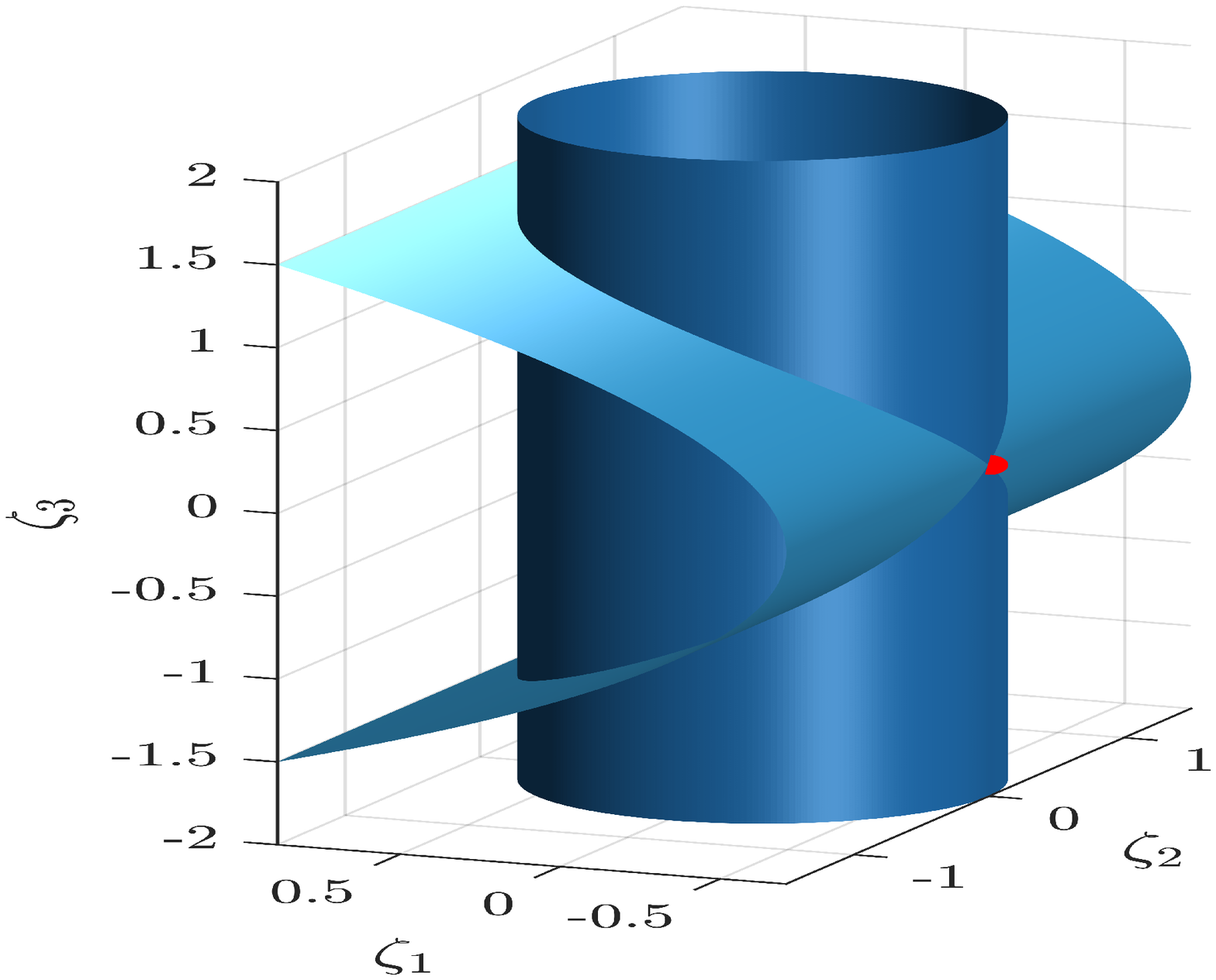}
\label{fig:energy_time_trajCV_z0}}
\subfloat[Optimal Trajectories $\alpha=30^o$]{
  \includegraphics[width=55mm,height=55mm]{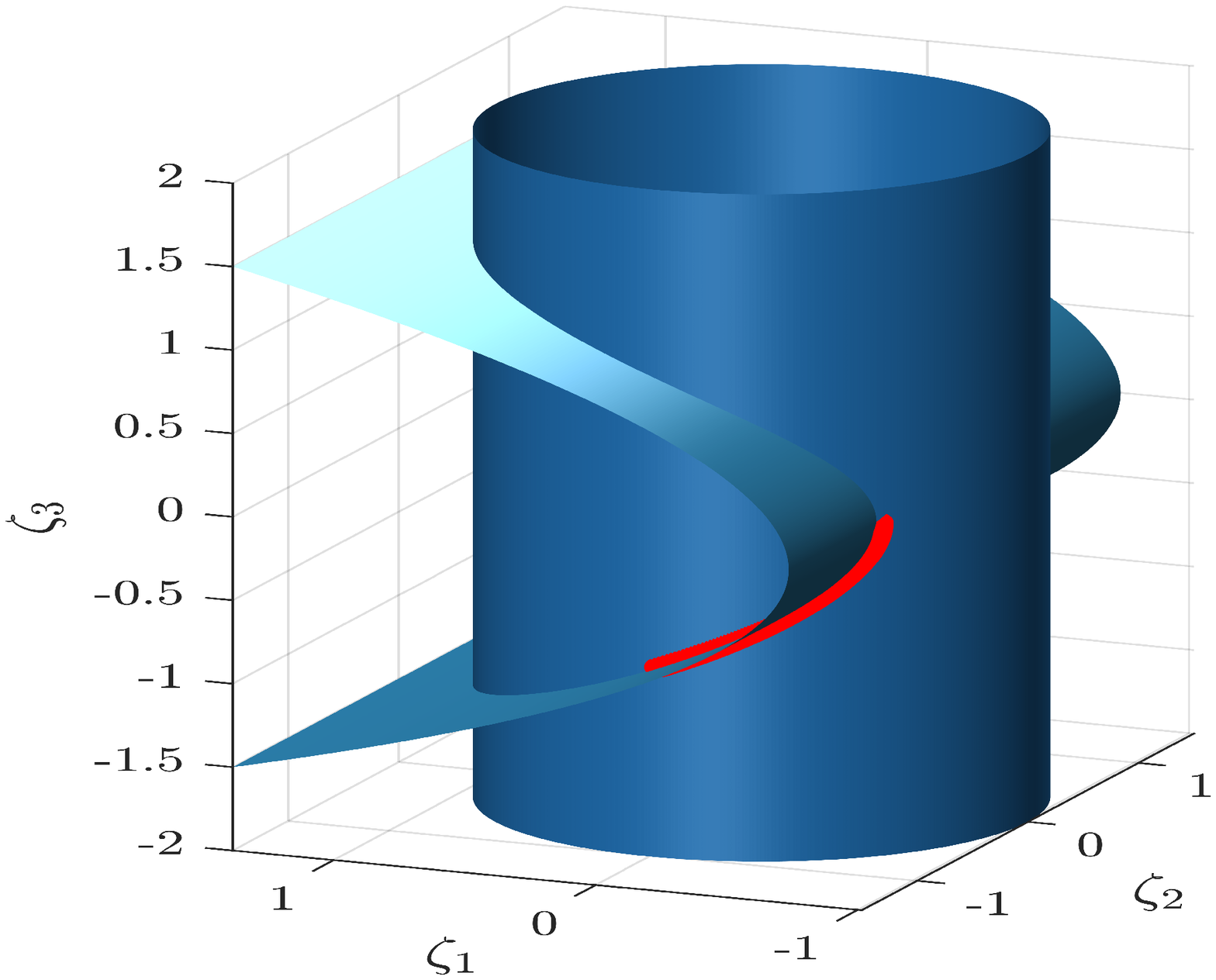}
\label{fig:energy_time_trajCV_z30}}
\hspace{0mm}
\subfloat[Optimal Trajectories $\alpha=60^o$]{
  \includegraphics[width=55mm,height=55mm]{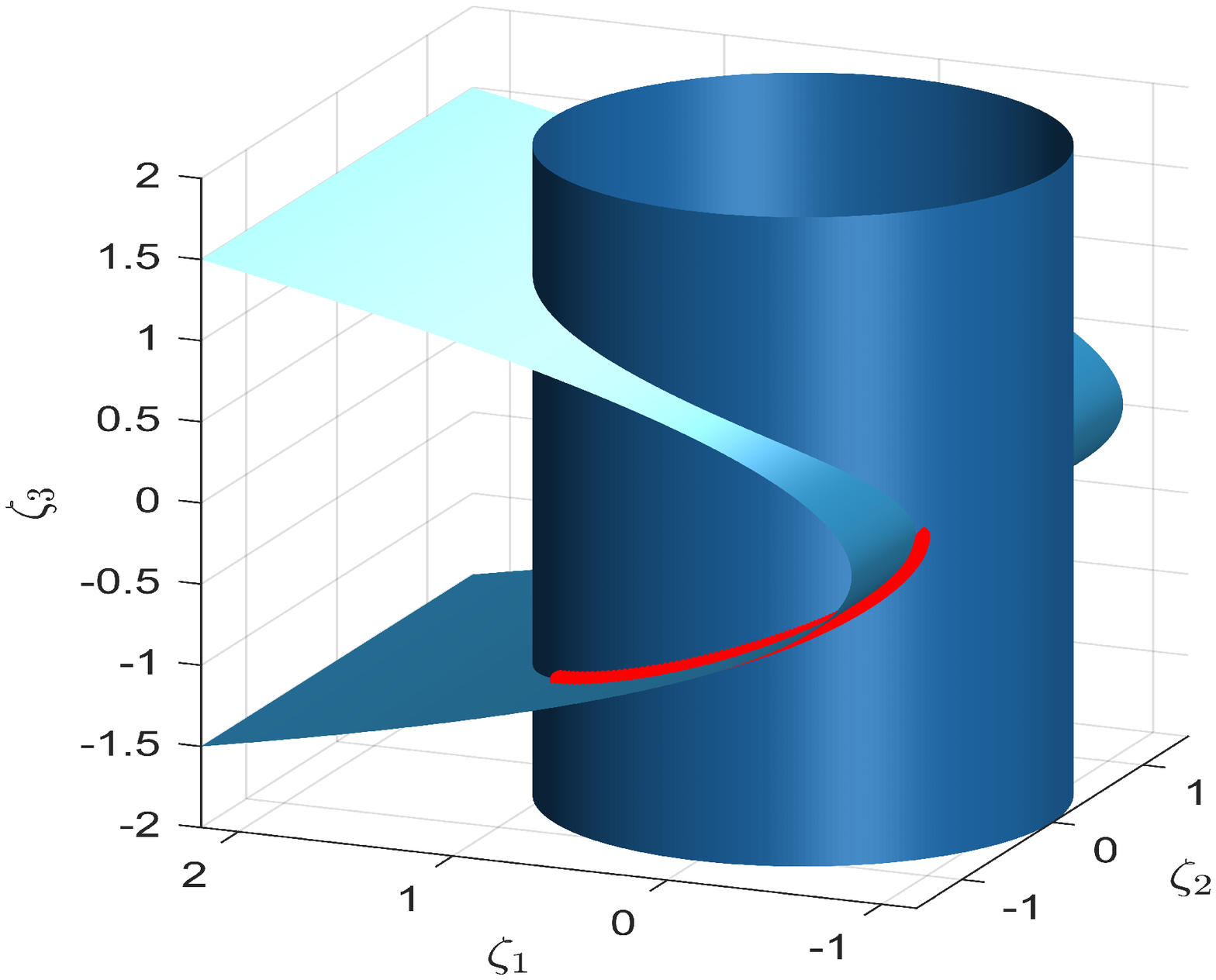}
\label{fig:energy_time_trajCV_z60}}
\subfloat[Optimal Trajectories $\alpha=90^o$]{
  \includegraphics[width=55mm,height=55mm]{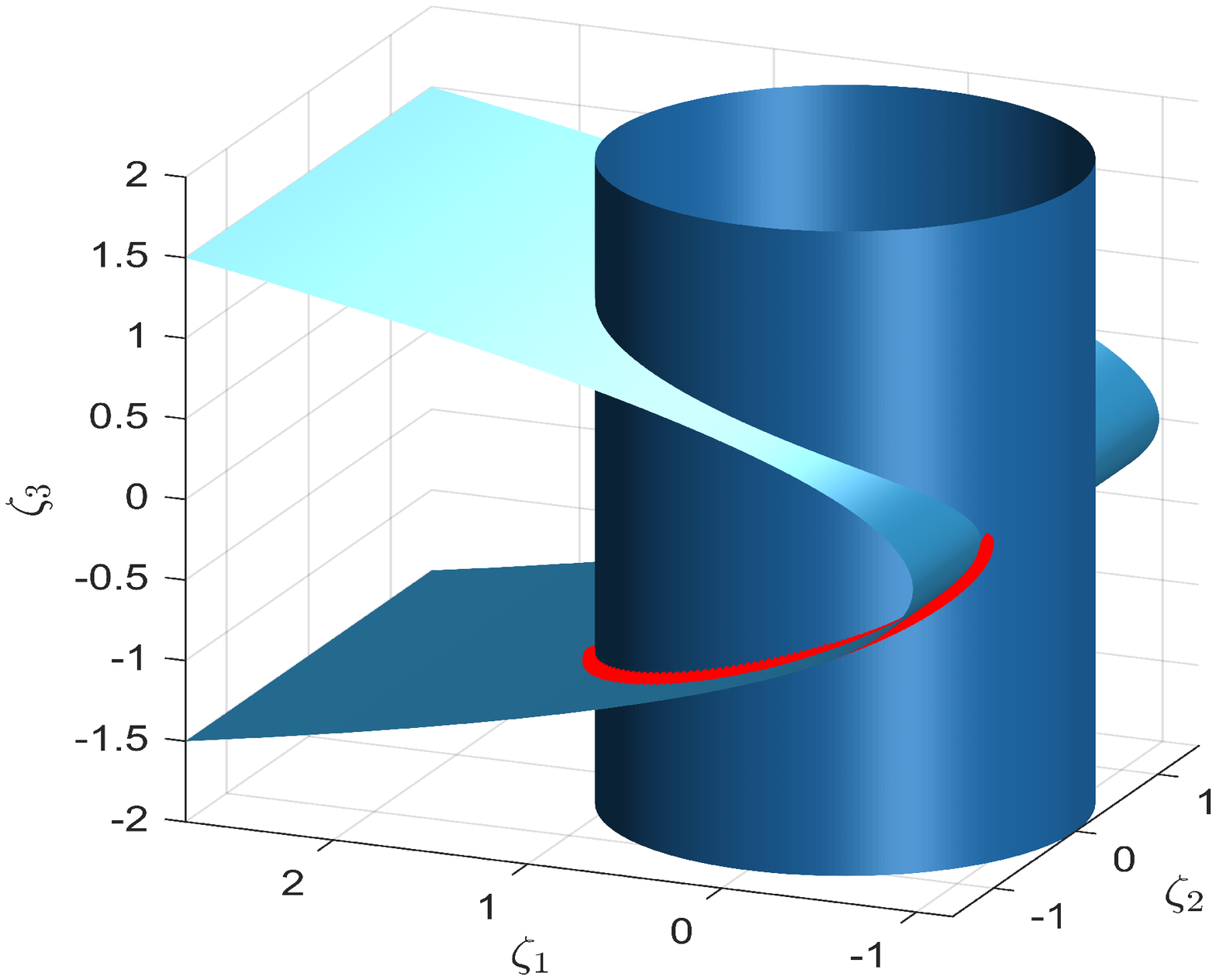}
\label{fig:energy_time_trajCV_z90}}
\caption{Optimal Costates for Fixed Linear velocity [R = 1 , $\mu =0.5$]}
\label{fig:opt_cos_cyl2}
\end{figure}

Figure~\ref{fig:opt_cos_cyl2} illustrate the evolution of the optimal costates for four distinct maneuvers which are seen to lie on the intersection of a cylinder and an extruded parabola.

It should be noted that the limiting case from Equation~\ref{eq:root_ineq} corresponds to:
\begin{equation}
    K_c = \left(\frac{C_c}{v_c \mu}\right)
\end{equation}
which results in the modulus $m=1$ which degenerates the Jacobi elliptic functions to hyperbolic functions. This leads to the optimal control $\omega$ resulting in:
\begin{equation}
\omega(\tau) = \mp\frac{\sqrt{2\mu v_c {K_c}+2C_c}}{\mu} \,\text{cn}(\Lambda\tau+\mathcal{K}(1) - \Lambda,1) = \mp\frac{\sqrt{2\mu v_c {K_c}+2C_c}}{\mu} \,\text{sech}(\Lambda\tau+\mathcal{K}(1) - \Lambda) = 0,
\end{equation}
which corresponds to the terminal constraint lying on a line connected to the origin whose angle is coincident with the initial orientation of the WMR. This case is illustrated geometrically in the $\zeta_1-\zeta_2-\zeta_3$ space in Figure~\ref{fig:energy_time_trajCV_z0}. 
The corresponding optimal costates are all constant and are given by:
\begin{align}
    &\omega(\tau) = 0, & \theta(\tau) = 0, && v(\tau) = \frac{\sqrt{2\mu(1-\mu)}}{\mu}, &\\
    & x(\tau) = \frac{\sqrt{2\mu(1-\mu)}}{\mu} \tau, & y(\tau) = 0, && T_f = \frac{\mu}{\sqrt{2\mu(1-\mu)}}, & \\
    & \zeta_1(\tau) = -\sqrt{2\mu(1-\mu)}, & \zeta_2(\tau) = 0, &&\zeta_3(\tau) = 0. & 
\end{align}
It should be noted that the optimal control for the limiting case (terminal state lying on the $x$-axis) is coincident for the two problems considered in this work, one where the linear velocity is unconstrained and the other where it is constrained to be the same over the entire maneuver.

\section{Conclusion}

This paper considers the problem of minimizing a weighted combination of energy and maneuver time for a wheeled mobile robot undergoing point-to-point maneuvers in an obstacle free space. Closed form solutions for the optimal control are derived which are functions of Jacobi elliptic functions parameterized by variables which are identified by solving simultaneous nonlinear equations. Two classes of problems are considered: (1) where the linear and angular velocities are unconstrained and (2) where the linear velocity is constant, but whose magnitude is solved as part of the control design. Two kinematic models of the wheeled mobile robot that are related by a passive transformation are analyzed and geometric interpretation of the constraint on the costates are revealed. For the problem where the control inputs are unconstrained, it is shown that the controls have to lie on the surface of a cylinder where time evolves along the cylinder's axis. Counter intuitive results where the robot initially backs before moving towards the terminal states is demonstrated for some regions of the operating space. Future work will use the closed form solution for the evolving states to reduces the path planning problem in the presence of obstacles to a way point determination problem.
\clearpage

\renewcommand{\theequation}{A.\arabic{equation}}

\setcounter{equation}{0}

\section*{Appendix A : Derivation of closed-form expressions for general point to point maneuver}
Equation~\ref{eq:thdot_diff} is a nonlinear differential equation in $\theta(t)$:
\begin{align}\label{eq:theta^2}
\frac{1}{2}\dot{\theta}^2  = \frac{T_f^2z^2}{4\mu^2}\Big(Q- 2\sin^2(\theta+\phi)\Big)
\end{align}
where $Q = 1+c_\theta$. Define $\psi= \theta +\phi$ whose time derivative is $\dot{\psi} = \dot{\theta}$ since $\phi$ is a constant. Equation~\ref{eq:theta^2} can be rewritten as: 
\begin{align}\label{psi}
\dot{\psi}^2 =\frac{T_f^2z^2}{2\mu^2}\Big(Q- 2\sin^2(\psi)\Big).
\end{align}
Define $\Xi(\tau) = \sin(\psi(\tau))$ and the square of its derivative leads to the equation: 
\begin{align}\label{psi_2}
(\frac{d\Xi}{d\tau})^2 &=  \Big(1-\Xi^2\Big)\dot{\psi}^2 \nonumber \\
\Rightarrow \dot{\psi}^2& = \frac{1}{(1-\Xi^2)}(\frac{d\Xi}{d\tau})^2.
\end{align}
Equating Equations~\ref{psi} and \ref{psi_2} leads to: 
\begin{align}\label{xi_1}
\frac{1}{(1-\Xi^2)}(\frac{d\Xi}{dt})^2 &= \frac{T_f^2z^2}{2\mu^2}(Q- 2\Xi^2) \nonumber \\
\Rightarrow (\frac{d\Xi}{dt})^2 &= \frac{T_f^2z^2}{\mu^2}\Xi^4-\frac{T_f^2z^2}{\mu^2}(1+\frac{Q}{2})\Xi^2+\frac{QT_f^2z^2}{2\mu^2}.
\end{align}
Consider the solution of  $\Xi = A \: \text{sn}(b\tau+\eta,m) $, whose time derivative squared leads to:  
\begin{align}\label{xi_2}
(\frac{d\Xi}{dt})^2 &= A^2b^2\text{cn}^2(bt+\eta,m)\text{\text{dn}}^2(bt+\eta,m) \nonumber\\ 
&= A^2b^2(1-\text{sn}^2(bt+\eta,m))(1-m\text{sn}^2(bt+
\eta,m))\nonumber\\
& =\frac{b^2m}{A^2}\Xi^4-b^2(1+m)\Xi^2+A^2b^2.
\end{align}
We now have two $4^{th}$ order polynomial functions (right hand sides of Equation~\ref{xi_1} and \ref{xi_2}), where the unknown parameters $A$, $b$ and $m$ can be determined by matching the coefficient of Equation \ref{xi_1} and \ref{xi_2}. This leads the closed-form expression of $\theta(\tau)$:
\begin{equation}\label{eq_apx:CF_theta}
\tcbhighmath[drop fuzzy shadow]{\theta (\tau) = \arcsin\Bigg(\sqrt{\frac{Q}{2}}\text{sn}(\frac{T_fz}{\mu}\tau+\eta,\frac{Q}{2})\Bigg)-\phi}
\end{equation}
which is given in equation \ref{eq:theclf} in the main document, and the initial condition leads to the expression for $\phi$.
Using Equation~\ref{eq_apx:CF_theta}, the closed-from expression of the position states $x$ and $y$ can be derived. 
Since we have shown that (Equation \ref{eq_v}): 
\begin{align}
v = \frac{-z}{\mu}\sin(\theta+\phi),
\end{align}
the rate of change of position $x$ and $y$ (i.e $\dot{x}$ and $\dot{y}$) can be rewritten as: 
\begin{align}\label{xdot}
\dot{x}(\tau) & = -\frac{T_fz}{\mu}\sin(\theta+\phi)\cos(\theta)
\end{align}
\begin{align}\label{ydot}
\dot{y}(\tau) & = -\frac{T_fz}{\mu}\sin(\theta+\phi)\sin(\theta).
\end{align}
Substitute Equation \ref{eq_apx:CF_theta} into \ref{xdot} and \ref{ydot} leads to: 
\begin{align}
\dot{x}(\tau) & = -\frac{T_fz\sqrt{m}}{\mu}\text{sn}(u,m)\Big(\cos(\phi)\sqrt{1-m\,\text{sn}^2(u,m)}+\sin(\phi)\sqrt{m}\text{sn}(u,m)\Big)
\label{eq:xdot2}
\end{align}
\begin{align}
\dot{y}(\tau) & = -\frac{T_fz\sqrt{m}}{\mu}\text{sn}(u,m)\Big(\cos(\phi)\sqrt{m}\text{sn}(u,m)+\sin(\phi)\sqrt{1-m\,\text{sn}^2(u,m)}\Big)
\label{eq:ydot2}
\end{align}
Where $u = \frac{T_fz}{\mu}\tau+\eta $ and $m =\frac{Q}{2}$. Equations~\ref{eq:xdot2} and \ref{eq:ydot2} can be expanded and simplified as:
\begin{align}
\dot{x}(\tau) & =-\frac{T_fz\sqrt{m}}{\mu}\,\cos(\phi)\text{sn}(u,m)\text{dn}(u,m)-\frac{T_fzm}{\mu}\sin(\phi)\text{sn}^2(u,m)
\end{align}
\begin{align}
\dot{y}(\tau) & =-\frac{T_fzm}{\mu}\,\cos(\phi)\text{sn}^2(u,m)+\frac{T_fz\sqrt{m}}{\mu}\sin(\phi)\text{sn}(u,m)\text{dn}(u,m)
\end{align}
which can be integrated and the closed-form expressions for the position states $x$ and $y$ are: 
\begin{equation}\label{eq:cf_x}
\tcbhighmath[drop fuzzy shadow]{x(t) = \sqrt{m}\cos(\phi)\text{cn}(u, m)-\sin(\phi)(u-E(\text{amp}(u,m),m))+C_x}
\end{equation}
\begin{equation}\label{eq:cf_y}
\tcbhighmath[drop fuzzy shadow]{y(t) = -\sqrt{m}\sin(\phi)\text{cn}(u, m)-\cos(\phi)(u-E(\text{amp}(u,m),m))+C_y}
\end{equation} 
which verify Equation \ref{eq:const1} and \ref{eq:const2}. Similarly, the initial condition leads to the expressions for $C_x$ and $C_y$.

Similarly, since $\dot{\lambda}_3(\tau) = -T_fz\,v\cos(\theta+\phi)$, it can be rewritten in terms of Jacobi elliptic functions as: 
\begin{align}\label{eq:lam3_dot}
\dot{\lambda}_3  = \frac{T_fz^2}{\mu}\sin(\theta+\phi)\cos(\theta+\phi) = \frac{T_fz^2\sqrt{m}}{\mu}\text{sn}(u,m)\text{dn}(u,m)
\end{align}
and can be integrated resulting in the closed-form expression for the third costate:
\begin{equation}\label{eq:cf_lam3}
\tcbhighmath[drop fuzzy shadow]{\lambda_3(\tau) = -z\sqrt{m}\text{cn}(u,m)+C_{\lambda_3}}
\end{equation}
which is shown in equation \ref{eq:lam3closed}. It should be noted that equations \ref{eq:clam3} to \ref{eq:clam34} were used to demonstrate that the integration constant $C_{\lambda_3} = 0$.

Lastly, the closed-from solution of the fourth costate can be rewritten in terms of only $v$ and $\omega$ by using the necessary conditions from equation \ref{eq:L4dot}, \ref{eq:optv}, and \ref{eq:optomg}:
\begin{align}\label{eq:lambda_4}
\dot{\lambda_4} = -1+\mu+\frac{\mu}{2}v^2 + \frac{\mu}{2}\omega^2.
\end{align}
Since $v = \frac{-z}{\mu}sin(\theta+\phi)$, the closed-form solution for $v$ can be obtained using equation~\ref{eq_apx:CF_theta}:
\begin{align}
v(\tau) & = -\frac{z}{\mu}\sin\big(\arcsin\big(\sqrt{m}\text{sn}(u,m)\big)\big) = -\frac{z\sqrt{m}}{\mu}\text{sn}(u,m)
\end{align}
and $\omega$ is $\frac{-\lambda_3}{\mu}$, therefore, the analytical expression is: 
\begin{align}
\omega(\tau) = \frac{z\sqrt{m}}{\mu}\text{cn}(u,m)
\end{align}
which can be substituted into equation~\ref{eq:lambda_4} and it can be rewritten as:
\begin{align}
\dot{\lambda_4} = -1+\mu+\frac{z^2m}{2\mu}
\end{align}
and the variables $\mu$, $z$ and $m$ are constant parameters. The closed-form expression of the fourth costate is: 
\begin{align}\label{eq:cf_lam4}
\tcbhighmath[drop fuzzy shadow]{\lambda_4 = (-1+\mu+\frac{z^2m}{2\mu})\tau +C_{\lambda_4}}.
\end{align}
The transversality condition leads to $C_{\lambda_4} =0$ which is proven in equation \ref{eq:lam4cf} in the main manuscript.	
\clearpage

\setcounter{table}{0}
\renewcommand{\thetable}{B. \arabic{table}}

\setcounter{figure}{0}
\renewcommand{\thefigure}{B. \arabic{figure}}
\section*{Appendix B : Computation Cost}\label{sec:App_B}
	To address the computational cost and show the benefit of the proposed approach, a series of numerical simulations were conducted. The  Tic-Toc commanded from MATLAB R2020b was used to measure the computational time for the algorithm to converge. 
	
	 The proposed approach leads to analytical expressions which permits reformulating the optimal control problem as a nonlinear programming problem.
		To evaluate the computational time required for convergence, first, we choose the terminal location arbitrarily, ($[x_f,y_f] = [cos(30), sin(30)]$), and the initial guesses for the unknown variables $Q$ and $T_f$ are perturbed uniformly around the optimal ($[Q^*,T_f^*] = [1.21, 0.94)]$). Table~\ref{table:NLP} presents the time required for convergence for entire testing cases. It can be seen that all initial guesses converge to the optimal solution.
	\begin{table}[htbp] 
	\centering
	\vspace{-2pt}
	\small
	\begin{tabular}{|c||c|c|c|c|c|}
		\hline
		\backslashbox{$T_f$}{$Q$} & 1.19 & 1.20 & 1.21 & 1.22 & 1.23 \\
		\hline
		\hline
		0.92 & 3.173 &	1.911 &	1.907 &	1.781 &	1.938 \\
		\hline
		0.93 &	1.792 &	1.771 &	1.780 &	1.779 &	1.808 \\
		\hline
		0.94 &	1.803 &	1.774 &	1.426 &	1.781 &	1.767 \\
		\hline
		0.95 &	1.777 &	1.798 &	1.769 &	1.794 &	1.791 \\
		\hline
		0.96 &	1.767 &	1.806 &	1.792 &	1.782 &	1.798
	  \\ \hline
	\end{tabular}
	\caption{Computation Cost and Convergence of Nonlinear Programming Problem}
		\label{table:NLP}
	\vspace{-7pt}
\end{table}

To compare, the given optimal control problem is solved numerically using the shooting method, where it requires four unknown variables $\lambda_1$, $\lambda_2$, $\lambda_3$, and $T_f$ to be solved. The same terminal conditions are used, and eleven different initial guesses for each unknown variables are randomly selected around its optimal and  [$\lambda_1^*$,$\lambda_2^*$,$\lambda_3^*$,$T_f^*$]  = [-0.17,-0.89,-0.68,0.94] leads to total of $11^4$ testing scenarios are examined. Convergence statistics of the shooting method revealed an anemic 1\% convergence rate to the optimal solution. 



\clearpage 
\renewcommand{\theequation}{C.\arabic{equation}}

\setcounter{equation}{0}

\section*{Appendix C : Derivation of closed-form expressions for fixed Linear Velocity}
From equation \ref{eq:vcont_cost} to \ref{eq:zeta5_closed} we have demonstrated the derivation of the closed-form expressions of $\zeta_1$, $\zeta_2$, $\zeta_3$, $\zeta_5$ and $\omega$ for the fixed linear velocity case. Using the information we now can derive the closed-form solutions of $\theta$, $x$ and $y$. 

First, the closed-form solution of orientation $\theta$ can be obtained by using the analytical solution of $\omega$. Since $\omega=\frac{\sqrt{2\mu v_c {K_c}+2C_c}}{\mu} \,cn(\Lambda\tau+\eta,m)$, the time derivative of $\theta$ can be written as: 
\begin{align}\label{eq_apx:dubin_thdot}
\dot{\theta} = T_f\omega = \frac{T_f\sqrt{2\mu v_c {K_c}+2C_c}}{\mu}\text{cn}(\Lambda\tau+\eta,m)
\end{align}
which can be integrated and the solution is: 
\begin{align}
\theta(\tau) = \frac{T_f\sqrt{2\mu v_c {K_c}+2C_c}}{\Lambda\mu\sqrt{m}}\arccos(\text{dn}(\Lambda\tau+\eta,m))+C_\theta,
\end{align}
where $m = \frac{{K_v} v_c \mu+C_c}{2{K_c} v_c \mu}$ and $\Lambda =\sqrt{\frac{v_cT_f^2}{\mu}{K_c}}$.
Replace $\Lambda$ and $m$ and simplifying leads to:  
\begin{equation}\label{eq:dubin_cf_theta}
\tcbhighmath[drop fuzzy shadow]{
	\theta(\tau) = 2\arccos(\text{dn}(\Lambda\tau+\eta,m))+C_\theta}
\end{equation}
Using equation~\ref{eq:dubin_cf_theta}, the closed-from expression of position state $x$ and $y$ can be derived.The rate of change of position $x$ and $y$ (i.e $\dot{x}$ and $\dot{y}$) can be rewritten as:
\begin{align}\label{eq:dubin_xdot}
\dot{x}(\tau) &= T_fv_c\cos(\theta)  =  T_f v_c \cos(2 \arccos(\text{dn}(u,m))+C_\theta)
\end{align}
\begin{align}\label{eq:dubin_ydot}
\dot{y}(\tau) &= T_fv_c\sin(\theta)  =  T_f v_c \sin(2 \arccos(\text{dn}(u,m))+C_\theta).
\end{align}
Using trigonometric identities, Equations~\ref{eq:dubin_xdot} and \ref{eq:dubin_ydot} can be rewritten as:
\begin{align}
\dot{x}(\tau) & =  T_f v_c \Biggr((2\text{dn}^2(u,m)-1)\cos(C_\theta)-2\sqrt{m}\text{sn}(u,m)\text{dn}(u,m)\sin(C_\theta)\Biggr)
\end{align}
\begin{align}
\dot{y}(\tau)& =  T_f v_c \Biggr(2\sqrt{m}\text{sn}(u,m)\text{dn}(u,m)\cos(C_\theta)+(2\text{dn}^2(u,m)-1)\sin(C_\theta)\Biggr),
\end{align}
which can be integrated and the closed-form equations for $x$ and $y$ are:
{\small
\begin{equation}
\tcbhighmath[drop fuzzy shadow]{
	x(\tau) = T_f v_c \Biggr(\frac{2\cos(C_\theta)}{\Lambda}E(amp(u,m),m)-\cos(C_\theta)\tau+\frac{2\sqrt{m}\sin(
		C_\theta)}{\Lambda}\text{cn}(u,m) \Biggr)+C_x\label{eq:x_closed}}
\end{equation}
\begin{equation}
\tcbhighmath[drop fuzzy shadow]{
	y(\tau) = T_f v_c \Biggr(\frac{2\sin(C_\theta)}{\Lambda}E(amp(u,m),m)-\sin(C_\theta)\tau-\frac{2\sqrt{m}\cos(C_\theta)}{\Lambda}\text{cn}(u,m) \Biggr)+C_y. \label{eq:y_closed}}
\end{equation}}The integration constants $C_x$ and $C_y$ can be obtained by using the initial condition.

\bibliography{ref}

\end{document}